\documentclass[a4paper,oneside,bibliography=totoc]{scrbook}

\usepackage[utf8]{inputenc}
\usepackage[T1]{fontenc}
\usepackage{lmodern}
\usepackage{longtable}
\usepackage{graphicx}
\usepackage{latexsym}
\usepackage{amsmath}
\usepackage{amssymb}
\usepackage{tabularx}
\usepackage{booktabs}
\usepackage{float}
\usepackage{array} 
\usepackage{caption}
\usepackage[numbers]{natbib}
\usepackage{algorithm} 
\counterwithin{algorithm}{chapter} 
\usepackage{algorithmic}
\usepackage{csquotes}
\usepackage{url}

\usepackage[dvipsnames]{xcolor}
\usepackage[colorlinks,citecolor=Green]{hyperref} 
\usepackage{lipsum} 

\usepackage{natbib}
\setcitestyle{authoryear,round,semicolon,aysep={},yysep={,}} \let\cite\citep

\begin{document}

\frontmatter \subject{Preprint} 
\title{Predicting Length of Stay in Neurological ICU Patients Using Classical Machine Learning and Neural Network Models: A Benchmark Study on MIMIC-IV}
\author{Alexander Gabitashvili (1), Philipp Kellmeyer (1), (2) \\
  } \date{May 23, 2025}
\publishers{ (1) Data and Web Science Group\\
		School of Business Informatics and Mathematics\\
		University of Mannheim\\
		Contact: philipp.kellmeyer@uni-mannheim.de\\		
		(2) Human-Technology Interaction Lab\\
		     Department of Neurosurgery\\
		     University of Freiburg - Medical Center}
\maketitle

\chapter{Abstract}

Intensive care unit (ICU) is a crucial hospital department that handles life-threatening cases. Nowadays machine learning (ML) is being leveraged in healthcare ubiquitously. In recent years, management of ICU became one of the most significant parts of the hospital functionality (largely but not only due to the worldwide COVID-19 pandemic).

This study explores multiple ML approaches for predicting LOS in ICU specifically for the patients with neurological diseases based on the MIMIC-IV dataset. The evaluated models include classic ML algorithms (K-Nearest Neighbors, Random Forest, XGBoost and CatBoost) and Neural Networks (LSTM, BERT and Temporal Fusion Transformer). 

Given that LOS prediction is often framed as a classification task, this study categorizes LOS into three groups: less than two days, less than a week, and a week or more. As the first ML-based approach targeting LOS prediction for neurological disorder patients, this study does not aim to outperform existing methods but rather to assess their effectiveness in this specific context. The findings provide insights into the applicability of ML techniques for improving ICU resource management and patient care.

According to the results, Random Forest model proved to outperform others on static, achieving an accuracy of 0.68, a precision of 0.68, a recall of 0.68, and F1-score of 0.67. While BERT model outperformed LSTM model on time-series data with an accuracy of 0.80, a precision of 0.80, a recall of 0.80 and F1-score 0.80.

\begingroup%
\hypersetup{hidelinks}
\tableofcontents%
\endgroup


\mainmatter

\chapter{Introduction}
\label{ch:intro}

\section{Background and Context}
Intensive Care Units (ICUs) are specialized hospital departments designed to provide intensive care for critically ill patients, offering advanced medical support that goes beyond what is available in standard hospital circumstances. These units play a crucial role in modern healthcare systems, ensuring that patients with life-threatening conditions receive constant monitoring, immediate medical intervention, and individual treatment plans to improve survival rates and overall recovery. 

Although the concept of an ICU is rather relative: its capacity and functions varying significantly across different healthcare systems, certain core principles identify its operations. These variations are influenced by factors such as available resources, hospital infrastructure, and the medical care approaches adopted in different regions. However, a key defining element of an ICU is its physical space, which is typically designed to accommodate specialized medical equipment, facilitate continuous patient monitoring, and support advanced treatments such as mechanical ventilation, hemodynamic stabilization and extracorporeal membrane oxygenation (ECMO). The layout of ICUs is often planned to optimize efficiency, ensuring that healthcare professionals can rapidly respond to critical situations while maintaining infection control protocols. 

ICUs also provide multidisciplinary care, involving a diverse team of healthcare professionals, including nurses, respiratory therapists, nutritionists, pharmacists, and rehabilitation specialists, all working together to manage complex cases effectively. By working together all the mentioned experts deliver comprehensive care, while addressing not only the patient's immediate needs from the medical perspective, but also also by deciding on the long-term recovery strategies and plans. Due to the extreme conditions managed in the ICU, the role of nurses is of higher importance, as they provide monitoring, administer medications, and, naturally, respond to any signs of deterioration.
Generally, ICUs comprise of several units, which level of the support capabilities vary within the same hospital. Some of them specialize to address the conditions to the specific patient group. For example, trauma ICUs manage patients with severe injuries from accidents or falls, while neurological ICUs specialize in conditions such as stroke, traumatic brain injury, or post-surgical recovery from neurosurgical procedures. In a similar fashion, cardiac ICUs (CICUs) focus particularly on patients with heart-related conditions, including those who recover from cardiac surgery or experience heart failure. Another example of ICUs, Respiratory ICUs, treats patients suffering from severe respiratory failure, quite often requiring advanced ventilatory support for a patient. Different natures of these examples are designed to provide the appropriate treatment based on the patients' conditions. This is made to improve the treatment outcomes. 

Beyond individual patient care, ICUs also play an crucial role within the broader hospital system. Many ICUs provide consultations, with an aim at assisting in the early recognition of those patients who may require intensive care before their condition deteriorates further. Additionally, ICU teams are often involved in post-discharge follow-up checks to monitor patients' long-term recovery. This is done to ensure that the survivors of critical illnesses do not have long-term negative effects on physical and cognitive function. ICU patients frequently require extended rehabilitation even after the discharge to regain their strength and movement independence. 

Moreover, the interdisciplinary nature of ICU teams in combination with recently developed state-of-the-art monitoring technologies are crucial for managing life-threatening conditions effectively. The usage of both noninvasive and invasive monitoring techniques allows for precise decision-making and timely interventions. Consequently, ICUs also play a significant role the medical research and education of the practitioners, while also undergoing quality improvement initiatives. Signficant share of the ICUs participate in clinical trials, investigating new treatment strategies and technologies to enhance patient outcomes. ICUs are an often used as a setup for the education of medical residents, nurses, and other healthcare professionals in the principles of critical care \cite{marshall2017}.

\section{Problem Statement}
Having reliable estimation for the remaining LOS of patients in ICU is essential for the resource planning of hospital: with such solution resources can be utilized more efficiently and provide insights in advance, leading to better patient care.

Although many researches have been conducted to assess ability of machine learning to predict LOS for patients in ICU, there is still a gap in using the domain knowledge and ML techniques specifically for patients with neurological disorders. Therefore, in order to provide the best possible results a study must take into account particular challenges and circumstances of such group of patients, which result in unique feature generation and, consequently, more informed feature selection.

\section{Research Objectives}
In order to provide hospitals with a suitable solution for predicting LOS in ICU for patients with neurological disorders following key research questions are selected:

\begin{itemize}
\item Which ML models perform best for LOS prediction in neurological patients?
\item Which data preparation techniques result in higher score on the selected metrics?
\item How does performance vary across different LOS categories (e.g., less than 2 days, less than 1 week, 1 week or more)?
\end{itemize}

To prepare and test proposed models recently updated Medical Information Mart for Intensive Care Dataset (MIMIC-IV) will be used. \cite{johnson2024mimic}

Overall, while current research provides an overview of different models' performance for ICU length of stay (LOS) prediction, it also presents a comprehensive pipeline, from data extraction from the MIMIC dataset to model training and evaluating.

\section{Methodology Overview}

Prior to model training, an \textbf{Exploratory Data Analysis (EDA)} will be conducted to gain insights into the dataset, focusing on patients diagnosed with \textbf{neurological disorders}. This step will help wth identifying patterns, detecting anomalies, and helping during feature selection step.

The final modeling pipeline will be structured as follows:

\begin{enumerate}
    \item \textbf{Data Retrieval:} \\
    Loading the preprocessed data from storage, including additional \textbf{column transformations} where necessary (e.g., encoding categorical variables, scaling continuous variables).

    \item \textbf{Feature Selection:} \\
    Employing techniques such as model-wise \textbf{feature selection}, \textbf{correlation analysis}, and \textbf{domain knowledge} to identify the most predictive features.

    \item \textbf{Hyperparameter Optimization:} \\
    Leveraging optimization techniques such as \textbf{Optuna} framework to fine-tune model parameters for their optimal performance.

    \item \textbf{Model Training and Evaluation:} \\
    Training selected models and assessing their performance using appropriate metrics (most popular ones from the related work), with cross-validation to ensure generalizability.

    \item \textbf{Feature Importance Analysis:} \\
    Analyzing independent feature contributions using \textbf{feature importance scores} from treebased models to interpret model behavior and identify key factors influencing ICU LOS for classic ML algorithms and using Permutation Importance approach for LSTM and transformer models, as they do not offer out-of-the-box feature importance functionality.

    \item \textbf{Model Comparison and Validation:} \\
    Comparing traditional ML models against LSTM and Transformer-based models to assess their relative performance and generalizability.
\end{enumerate}

\chapter{Literature Review}

In this section related terms and papers will be discussed.

Firstly, the structure of MIMIC-IV dataset will be discussed and how it may help achieve aforementioned goals of the current paper. Then, works and papers solving similar problem will be reviewed and analyzed on how the employed techniques there might help with the research. After that, the relevant results from the papers will be analyzed in order to later compare with the results of the current research.

\section{Intensive Care Units and Workload Prediction}

To start with, we will assess how do Intensive Care Units (ICUs) function in order to understand the current practices in ensuring that all the needed resources are available and their workload can be predicted.

Tasks of nurses play a key role in the ICUs due to their responsibility for monitoring patient's conditions, managing ventilators and providing care in critical circumstances. The complexity and intensity of care required in ICUs lead to significant nursing workload, which directly impacts patient survivability and length of stay, while also the total efficiency of the unit. Therefore, accurately predicting and managing nursing workload is essential for maintaining high standards of critical care and preventing staff from overload.

For such purposes several methods have been developed and implemented in hospitals, so that help with planning the workload by estimating the length of stay (LOS) of the patients. These methods range from traditional clinical scoring systems to advanced machine learning models, each with varying degrees of applicability.

\begin{itemize}
    \item \textbf{APACHE (Acute Physiology and Chronic Health Evaluation)}: Initially designed to predict patient mortality, APACHE scores are also used to estimate resource utilization and nursing workload based on disease severity \cite{zimmerman2006apache}.
    \item \textbf{SOFA (Sequential Organ Failure Assessment)}: This score tracks organ dysfunction over time and helps predict the intensity of care required, indirectly influencing workload predictions \cite{raith2017sofa}.
    \item \textbf{NEMS (Nine Equivalents of Nursing Manpower Use Score)}: Specifically developed to quantify nursing workload, NEMS assigns scores based on the complexity of interventions required, such as mechanical ventilation or dialysis. It is widely used in Europe and Canada for real-time workload assessment and staffing decisions \cite{reis1997nems}.
\end{itemize}

Hospitals utilize Electronic Health Records (EHRs) and Hospital Information Systems (HIS) to analyze historical and real-time data for workload prediction. Key strategies include:

\begin{itemize}
    \item \textbf{Bed Occupancy Forecasting}: By analyzing admission and discharge patterns and seasonal trends hospitals can predict ICU bed demand and corresponding nursing workload.
    \item \textbf{Patient Flow Models}: These models assume and simulate patient movement between departments, such as from emergency rooms or operating theaters to the ICU, allowing for proactive workload management \cite{long2018boarding}.
    \item \textbf{Length of Stay (LOS) Prediction Models}: Predicting the length of ICU stays helps in estimating when beds will become available, helping in resource and staff planning.
\end{itemize}

Among these approaches, the prediction of LOS in ICUs stands out as the most critical. Accurate LOS predictions directly helps with understanding of the future workload assessments and bed availability. This assists in adapting a suitable nursing staff allocation and the distribution of patients. 

\section{Papers Focusing on the Predicting LOS in ICUs}
In recent years, there has been a significant rise of interest regarding the prediction of LOS in ICUs. This growing attention matches with the industry's need to optimize resource allocation, improve patient care, and enhance hospital management efficiency. As a result, several researchers have focused on developing a unified approach for LOS prediction that is applicable across various patient conditions.

The main objective of these studies has been to implement generalized models trained on the relevant datasets, that cover the heterogenous patients, admitted to ICUs. These models were designed to predict LOS without significantly relying on the specific medical conditions of individual patients. By implementing and developing models in that fashion, researchers have aimed to create widely applicable models capable, that are suitable to function across different ICU settings and patient demographics. However, there is also drawbacks associated with that approach, namely limitations associated with the overlooking the nuanced impact of particular diseases and unique patient trajectories on LOS.

Consequently, while these models offer higher applicability, they leave a significant gap for more specialized future research. Specialized studies focusing on specific patient groups, such as those with neurological disorders, cardiac conditions, or post-surgical recovery cases, are crucial for better assessment of patients' LOS in more targeted contexts. Addressing this gap could lead to more accurate and clinically relevant LOS forecasts. That contribution to more effective ICU management may improve patient outcomes.

\subsection{Prediction of Length-of-stay at Intensive Care Unit (ICU) Using Machine Learning based on MIMIC-III Database, \textit{Hasan et al.}}

\textit{Hasan et al.} conducted their research using a previous version of the MIMIC dataset, MIMIC-III, to develop predictive models for LOS in ICU for the patients regardless of their disease. The study aimed to assess the effectiveness of different machine learning algorithms for regression-based LOS prediction.

To achieve this, the researchers experimented with four different algorithms: Support Vector Machine (SVR), Random Forest (RF), XGBoost, and Voting Regressor (VR). The Voting Regressor functioned as an ensemble model, aggregating the predictions from XGBoost, Random Forest, and Support Vector Machine to produce an averaged output. By leveraging multiple learning techniques, they sought to enhance prediction accuracy and model robustness.

Among their key findings, the study showed that the XGBoost model outperformed all other tested approaches. It achieved the lowest Root Mean Squared Error (RMSE) of 1.2. This result indicates the qualities and potential of XGBoost in ICU-related predictive tasks, specifically for estimating patient LOS. The researchers emphasized the importance of ensemble learning techniques and feature selection in further improving model accuracy and reliability.

The study also acknowledged several limitations that may have impacted the results. One major constraint was the usage of the demo version of the dataset, which contained only 100 different patients, limiting the models from learning the patterns of different conditions better. This limited sample size might not have allowed to achieve the generalizability of the developed models and, which is also important, may not accurately reflect real-world ICU patient demographic groups. Additionally, the study considered only six features for prediction, which may have limited the models' ability to capture the complexity of LOS in ICU factors.

Another notable finding was that the best-performing model identified the "ICD-9 code" feature—representing the disease classification of a patient as one of the most significant predictors of LOS, which actually only specifies the diagnose system's version, while not exactly representing significant (patient-wise) information. This suggests that further research could be conducted by integrating the whole dataset, which will allow a more detailed analysis of disease-specific conditions. By expanding the dataset, future studies could significantly enhance the predictive power of the used models in the initial research and potentially offer more precise and condition-specific LOS estimations. \cite{hasan2023pred}.

\subsection{Prediction of Intensive Care Unit Length of Stay in the MIMIC-IV Dataset, \textit{Hempel et al.}}

Another study aimed to address the same problem as previous research but used a more recent version of the MIMIC dataset, MIMIC-IV version 2.1. In addition to the more recent dataset, the researchers took a different methodological method by approaching the problem as both a regression and a classification task. For classification, they chose a binary classification strategy, distinguishing between long-term and short-term ICU stays. They defined stays of fewer than five days as short-term and those that exceed five days as long-term.

To tackle these tasks, the researchers considered several machine learning algorithms, including Logistic/Linear Regression, Support Vector Machine (SVM), Random Forest, and XGBoost. Their findings indicated that for the classification task, the Random Forest model outperformed other algorithms, achieving an accuracy of 0.81 and an F1 score of 0.44. For the regression task, SVM yielded the best results in terms of Mean Absolute Error (MAE) at 1.68 days and Mean Absolute Percentage Error (MAPE) at 48.37 days. However, Random Forest achieved the best RMSE with a value of 2.81 days, indicating its superior performance in capturing LOS variations.

Despite these promising results, the study also had several limitations. One notable constraint was that the authors only considered patient data from the first 24 hours after ICU admission, which could have limited the models' ability to learn the changing patterns of patients' conditions. These learned patterns could have significantly helped in assessing the remaining LOS, allowing the real-time forecasting. Additionally, the study did not account for missing data, potentially affecting the accuracy and reliability of the results. To address these challenges, the authors suggested expanding the data collection window beyond the first 24 hours to improve the prediction of ICU stays exceeding four days, a refinement that could enhance future LOS prediction models \cite{hempel2023pred}.

\subsection{Predictors of in-hospital length of stay among cardiac patients: A machine learning approach, \textit{Daghistani et al.}}

Although the first two examples of related work aimed to provide a solution regardless of patients' diagnoses, there has been a growing interest in analyzing specific diseases and applying machine learning techniques to develop LOS prediction models designed with a focus on particular patient groups. One notable study, conducted by \textit{Daghistani et al.}, specialized on cardiac patients. Unlike many other studies that relied on publicly available datasets, this research utilized a dataset provided by the King Abdulaziz Cardiac Center (KACC), allowing for a more concrete approach.

In contrast to other research efforts that treated LOS prediction as a regression problem, the authors approached it as a solely classification task. They categorized each patient's total LOS into distinct classes based on the following classification scheme:

\begin{itemize}
    \item \textbf{Short}: 0–2 days
    \item \textbf{Medium}: 3–5 days
    \item \textbf{Long}: More than 5 days
\end{itemize}

This classification was determined using an equal-frequency binning approach, ensuring a balanced distribution of data across the classes. Regarding machine learning techniques, the researchers considered four different algorithms: Random Forest, Artificial Neural Network (ANN), Support Vector Machine (SVM), and Bayesian Network (BN). To enhance model performance and reduce dimensionality, a feature selection process was conducted, resulting in the selection of 21 relevant features.

Among the models tested, Random Forest outperformed the other algorithms by achieving an accuracy score of 0.8 and an F-score of 0.8. This shows the model’s strong predictive capabilities for classifying LOS of ICU stays of cardiac patients. Remarkably, one of the key strengths of this study was its rigorous feature selection process. Additionally, the research leveraged the full length of available historical data, which enhanced the reliability of its predictions and provided valuable insights for ICU management in cardiac care settings.

\subsection{Predictors of readmissions and length of stay for diabetes related patients, Alturki et al.}

Another research, concerning the development of an algorithm for predicting LOS in ICUs for the diabetes related patients, was conducted in 2019 at Qassim University. This study used a diverse dataset consisting of data, obtained from the \textbf{Electronic Heatlh Records (EHR)} of 130 US hospitals and a total of 101766 admissions, which is available at UCI Machine Learning Repository. 

Similar to the previously discussed papers, this one considered algorithms, such as Logistic Regression, Random Forest, Support Vector Machine and XGBoost. K-Nearest Neighbors (KNN) algorithm was also added to the list, although the researchers articulated that the feature importance extraction for that algorithm cannot be computed. 

Although the paper focused on two goals, predicting readmissions and LOS, only the latter is relevant for the current study. 
Researches also approached this task as a classification with the following classes:

\begin{itemize}
    \item short: (1–4 days)
    \item medium (5–8 days)
    \item long (9-14 days)
\end{itemize}

After a feature selection, only 25 features were left for the training. Regarding the results, SVM outperformed other models and achieved a score of 0.87 accuracy and 0.87 F1-score \cite{alturki2019pred}.

Another study, concerning the development of an algorithm for predicting LOS in ICUs specifically for patients with diabetes-related conditions, was conducted in 2019 at Qassim University, this research used a diverse dataset obtained from \textbf{Electronic Health Records (EHR)} across 130 US hospitals, consisting of a total of 101,766 admissions data. The dataset is publicly available through the UCI Machine Learning Repository.

Similar to the previously discussed studies, this research explored various machine learning algorithms, including Logistic Regression, Random Forest, Support Vector Machine (SVM), and XGBoost. Additionally, the researchers integrated the K-Nearest Neighbors (KNN) algorithm into their analysis. However, they noted a key limitation of KNN: feature importance extraction could not be extracted directly from the KNN model, which may have affected interpretability of the developed model.

While the study aimed at  two primary objectives: predicting both re-admissions and LOS, the latter one is more relevant to the current research. The authors approached LOS prediction as a classification problem, categorizing patient stays into the following three classes:

\begin{itemize}
    \item \textbf{Short}: 1–4 days
    \item \textbf{Medium}: 5–8 days
    \item \textbf{Long}: 9–14 days
\end{itemize}

To improve model performance, a feature selection process was conducted, reducing the number of input variables to 25.

Regarding model performance, the results demonstrated that the SVM model outperformed the other algorithms, achieving an accuracy score of 0.87 and an F1-score of 0.87. These findings highlighted the potential of SVM model for LOS prediction in diabetic patients, displaying effectiveness in handling complex relationships within the dataset \cite{alturki2019pred}.

\subsection{A Comparison of Supervised Machine Learning
Techniques for Predicting Short-Term In-Hospital
Length of Stay Among Diabetic Patients, Morton et al.}

A similar study on predicting LOS in ICU for the patients with diabetes was conducted five years earlier. The primary objective of this research was to compare the performance of four different machine learning algorithms in solving the same task: predicting LOS for diabetes patients in ICUs. The selected algorithms included Random Forest, SVM, Multi-Task Learning (MTL), and Multiple Linear Regression.

MTL is a machine learning approach that improves model performance by training several related tasks together. Instead of learning each task separately, MTL allows them to share knowledge, which helps improve generalization and learning efficiency \cite{caruana1997mtl}.

For training and evaluating the models, the researchers utilized the HCUP Nationwide Inpatient Sample database, which contains significant amount of patients' data, having in total more than 8 billion records. 

Among the evaluated algorithms, the SVM demonstrated the best performance. However, despite outperforming the other models, SVM still showed rather low predictive capabilities, achieving an accuracy score of 0.68 and an F1-score of 0.65. These results showed that while SVM may offer some advantages in predicting LOS for diabetic patients, further research, model-tuning and additional predictive features may be necessary to enhance overall performance \cite{morton2014pred}.

\section{Papers on Neurological Patients in ICUs}

The research on developing ML models to predict or estimate the LOS in ICUs for neurology patients remains and unexplored area. Neurological conditions often present unique challenges which can have a significant impact on the duration of ICU stays. This makes it essential to correctly identify the key factors of such conditions that contribute to the prolonged ICU stay.

In this regard, some studies have been conducted to identify the main indicators associated with the extended ICU stays for patients with the neurological disorders. These studies aim at assessing the medical and logistical challenges that healthcare practitioners face during the treatment of such patients. Major factors such as the severity of the neurological condition, the effectiveness of early interventions (if any), and the availability of specialized care all play a crucial role in determining patient outcomes. The findings of such papers are relatable to the current study, as they will provide a help not only in feature generation and feature selection for the end models, but also with the general idea of which steps are needed to be done in order to achieve better performance.

\subsection{How Does Care Differ for Neurological Patients Admitted
to a Neurocritical Care Unit Versus a General ICU? \textit{Kurtz et al.}}

For example, in their research \textit{Kurtz et al.} examined the difference in treating patients in regular ICUs and neuroitensive ones. 

The study analyzing 1,906 ICU patients found that 231 had a primary neurological diagnosis, with 22\% treated in a neuro-ICU and 78\% admitted to a general ICU. Patients in neuro-ICUs were more frequently transferred from outside hospitals and were more likely to suffer from hemorrhagic strokes, whereas ischemic strokes and traumatic brain injuries (TBIs) were more common in general ICUs.

In neuro-ICUs, patients underwent a higher number of invasive procedures, including tracheostomy, hemodynamic monitoring, and intracranial pressure (ICP) monitoring. Despite similar rates of EEG monitoring across both settings, neuro-ICU patients received fewer blood transfusions and IV sedation but had greater nutritional support. Additionally, while not statistically significant, do-not-resuscitate (DNR) orders were less frequent in neuro-ICUs, suggesting a potentially more aggressive treatment approach.

Previous research has indicated that neuro-ICUs improve survival and functional outcomes, though the mechanisms behind these benefits remain unclear. This study showed that neuro-ICU patients were more likely to receive invasive hemodynamic monitoring and early tracheostomy, both of which are linked to improved management of neurological injuries. Early tracheostomy may suggest quicker weaning from ventilation, which can potentially reduce the total ICU length of stay. Additionally, early tube feeding was more commonly implemented in neuro-ICUs, possibly to prevent malnutrition, though its long-term impact remains uncertain.

Assessment of the best practices suggests that neuro-ICUs often employ daily sedation breaks, which can reduce ventilation time and enable more accurate neurological tests. A more restrictive approach to blood transfusions was also observed, which aligns with an evidence that limiting transfusions can indeed improve survival rates in neurological patients. Moreover, the lower prevalence of DNR orders in neuro-ICUs may indicate a more proactive approach to intensive care.

However, the study had limitations. Conducted as a one-day snapshot in New York City, its findings may not be generalizable. The lack of standardized severity scoring, long-term outcome data, and detailed laboratory information further restricts the ability to draw definitive conclusions about care decisions.

Overall, neuro-ICUs appear to provide specialized and potentially more effective care for patients with neurological conditions through aggressive monitoring, early tracheostomy, and restrictive transfusion policies. Yet, further research is needed to confirm whether these strategies consistently lead to improved outcomes across various healthcare settings \cite{kurtz2011how}.

Based on this paper, for training ML models to predict LOS in neuro-ICU patients, a combination of neurological monitoring, hemodynamic and respiratory parameters, lab tests, sedation practices, nutrition, and admission characteristics should be considered. Including these features can improve model accuracy and better capture the complexities of neurocritical care.

\subsection{Factors Influencing Length of Stay in Neurosurgical Intensive Care Unit, Kumwilaisak et al.}

Another study, led by \textit{Kumwilaisak et al.}, aimed at determining the key factors influencing LOS in neurosurgical ICUs. The analysis of 178 admissions over three months revealed notable trends in ICU utilization. A majority of patients had short stays of less than three days, accounting for over three-quarters of admissions but only a third of total ICU days. Those staying between three and fourteen days, though fewer in number, consumed nearly half of the total ICU days. Prolonged stays more than fourteen days were uncommon but contributed disproportionately to overall ICU use. Among those with longer LOS, cerebral aneurysm was the most frequently observed diagnosis. The study identifies two primary contributors to extended ICU stays: triple-H therapy, which is used for aneurysmal subarachnoid hemorrhage, and mechanical ventilation. Other contributing factors included preoperative intubation, preoperative mechanical ventilation, and postoperative pulmonary complications. Based on that, the authors proposed that early tracheostomy and structured weaning protocols could significantly help reduce LOS. This is particularly applicable in the resource-limited environments.

While the study provides important insights into ICU resource utilization, it has several limitations. The data collection period was only three months. This not only limits the generalization of the proposed models, but also decreases the understanding of the seasonability of illness trends, which is crucial for any medical setup. The relatively small sample size, especially for patients with prolonged stays, limited the statistical significance and wide applicability of the findings. Additionally, the study does not integrate multivariate analysis, which makes it difficult to identify whether certain aspects (features) independently contribute to prolonged LOS. Moreover, while the research focuses on LOS, it does not examine the impact of prolonged stays on patient outcomes, such as mortality or neurological recovery. The exclusion of records with missing data also was not articulated enough, which might have resulted in a probable selection bias.

Despite these limitations, the study offers a relevant overview for managing NICU resources. By identifying mechanical ventilation and triple-H therapy as the key contributors to extended stays, it specifies certain cases where intervention could improve efficiency.

\subsection{Prediction of Length of Stay for Stroke Patients Using Artificial Neural Networks, Neto et al.}

While the paper "Prediction of Length of Stay for Stroke Patients Using Artificial Neural Networks" examines the use of artificial neural networks (ANNs) to predict the length of stay for stroke patients, it focuses on the total hospital stay rather than specifically on ICU stays of patients with a specific neurological disorder, stroke, which makes it highly relevant for the current study.

The research is motivated by the need for more accurate predictions that can aid in hospital resource management and improve patient care. To achieve this, the researchers define three different use cases to evaluate which attributes contribute most to the prediction process. These cases involve using the full dataset, a subset of 14 selected variables, and a further reduced subset of seven attributes.

The results indicate that the model trained with the smallest subset of features (third use case) achieved the lowest RMSE of approximately 5.9451, outperforming both the full dataset (6.2951 RMSE) and the 14-variable subset (7.6601 RMSE). This suggests that reducing the number of attributes can improve prediction accuracy, likely by minimizing noise in the data. It also indicates that a significant share of features are appear to be irrelevant for the LOS in ICU prediction. However, the second use case, which included 14 attributes, achieved the lowest MAE value of 4.5478, while the first and third cases had similar values of 4.6350 and 4.6354, respectively. Given the small difference in MAE values, the authors conclude that the third use case, selected using the CfsSubsetEval evaluator, is the optimal choice due to its superior RMSE performance.

This research highlights the potential of ANNs in medical decision-making, demonstrating how machine learning can assist in hospital planning and stroke patient management.

\section{MIMIC-IV Dataset} \label{sec:mimic}

The MIMIC-IV (Medical Information Mart for Intensive Care IV) dataset is a publicly available database containing de-identified health records of patients admitted to ICUs at the Beth Israel Deaconess Medical Center. It is an enhanced version of the earlier MIMIC-III dataset, including improvements in data structure, quality, and coverage.

MIMIC-IV is derived from two in-hospital database systems: a hospital-wide electronic health record (EHR) system and an ICU-specific clinical information system. Its creation followed three key steps:  

\begin{itemize}
    \item First, during the acquisition phase, data from patients admitted to the Beth Israel Deaconess Medical Center (BIDMC) emergency department or ICU between 2008 and 2022 were extracted. A relevant patients list was formulated to contain medical record tests of all relevant patients. After that, hospital source tables were then filtered to include only the relevant data to these patients. 
    \item Next, in the preparation phase, the data were reorganized to facilitate retrospective analysis. This steps included the denormalization of tables and restructurization of the data into a more appropriate format. Although, as a significant drawback, no data cleaning was performed to maintain the integrity. 
    \item Finally, in the deidentification process, all patient identifiers, as required by HIPAA regulations, were removed. Identifiers were replaced with randomly generated integers for patients, hospitalizations, and ICU stays. Structured data were filtered using lookup tables, and a free-text deidentification algorithm was applied where necessary to remove protected health information (PHI). Additionally, timestamps were shifted randomly by a fixed number of days per patient to preserve internal consistency while preventing direct temporal comparisons between different patients. 
\end{itemize}

After completing these steps, the dataset was exported in a comma-delimited format for use in research and analysis.

MIMIC-IV is organized into two primary modules: \textit{hosp} and \textit{icu}, reflecting their respective data sources. The \textit{hosp} module contains information extracted from the hospital-wide electronic health record (EHR), while the \textit{icu} module includes data from the ICU-specific clinical information system, MetaVision (iMDSoft). The dataset consists of 364,627 unique patients, with records covering 546,028 hospitalizations and 94,458 ICU stays.  

\subsection{hosp Module}  
The \textit{hosp} module includes detailed information on 546,028 hospitalizations for 223,452 individuals. It consists of patient demographics, admission details, intra-hospital transfers, laboratory tests, microbiology cultures, medication administration, prescriptions, provider orders, hospital billing, and service-related information. Some records extend beyond inpatient stays, including outpatient and emergency department data. The module also contains a \textit{provider table}, where deidentified provider IDs are contextualized within different data tables to track their role in patient care.  

All dates in MIMIC-IV have been shifted randomly to a future time period between 2100 and 2200 for deidentification purposes. Each patient is assigned an \textit{anchor year} and an \textit{anchor year group} to provide an approximate reference to their actual hospitalization period (e.g., "2011-2013"). The \textit{anchor age} ensures that patients over 89 years old are uniformly set to 91, preserving anonymity.  

The dataset also includes \textit{date of death information}, obtained from hospital and state records. To prevent inadvertent identification, only deaths occurring within one year of the last hospital discharge are recorded. If a patient survived beyond this period, the corresponding field remains empty.  

\subsection{icu Module}  
The \textit{icu} module, derived from the MetaVision system, documents patient stays in intensive care. The data is structured into a \textit{star schema}, linking ICU stays to various event tables, including intravenous and fluid inputs, outputs, procedures, timestamps, and other clinical observations. Each event table contains a \textit{stay\_id} to associate records with specific ICU patients and an \textit{itemid} to classify documented concepts. Additionally, caregiver data is stored in a separate table, where each provider is assigned a deidentified \textit{caregiver\_id}.  

The ICU module includes records for 94,458 ICU stays across 65,366 unique patients. An ICU stay is defined as a continuous sequence of transfers within an ICU, with consecutive movements within the same unit merged under a single \textit{stay\_id}. However, the ICU stays, that did not happen in a consequent fashion, were treated as different, even if a patient was readmitted after a planned procedure. 

MIMIC-IV provides a comprehensive and structured dataset for studying hospital and ICU patient trajectories, supporting advanced research in healthcare analytics and predictive modeling.

One of its key strengths is its granularity, allowing for detailed time-series analysis of patient trajectories, particularly for ICU patients. The dataset is widely used for research in predictive modeling, clinical decision support, and AI-driven healthcare innovations. However, challenges include its complexity, the need for domain expertise to interpret medical codes and timestamps, and potential biases due to the dataset being based on a single hospital system.

\section{Summary}

Literature overview of the current study shed the light on the most recent researches conducted in the setting of ICUs to predict LOS of patients. Although some of them aimed at developing a generalized models, which could work equally good for different group of patients, the more promising results are shown in the papers, which focused on the specific disorders, such as cardiac and diabetes. 

Nonetheless, there is still a research gap in the area of predicting LOS in ICU for the patients with neurological disorders. One of the most relatable papers proposed an ANN algorithm, however for predicting total LOS at a hospital for the stroke patients, providing valuable insights for this paper, while also not solving the current paper's objectives. Researches by \textit{Kumwilaisak et al.} and \textit{Kurtz et al.} also produced an overview of which features might be useful for predicting LOS in ICUs for the patients with neurological disorders. 

MIMIC-IV dataset, thanks to its structure, data volume, and recentness is rendered as the most prominent dataset for conducting such research. Its data covers all the needed use cases in order to succeed with the objectives of the current study.

The literature review conducted in this study provides a comprehensive examination of the latest research efforts focused on predicting LOS in ICUs. A significant part of work has explored various machine learning models designed to estimate patient stay durations, with some studies attempting to develop generalized models, applicable across different patient groups. However, while these generalized approaches offer broad applicability, their predictive performance often falls short when compared to models designed to specific medical conditions. Studies focusing on particular disorders, such as cardiac conditions or diabetes, have demonstrated more significant results, which indicates the importance of condition-specific model development.  

Despite these advancements, a notable research gap remains in the area of LOS prediction for ICU patients with neurological disorders. Unlike other conditions that have received more extensive attention, neurological cases remain understudied in the context of ICU-specific LOS predictions. One of the most relevant studies in this area proposed an ANN model for predicting the total LOS at a hospital for stroke patients. While this research provides valuable insights into the broader topic of stroke patient management, it does not fully address the specific objective of this study, which is to predict ICU LOS rather than full hospital stays.  

Additional studies, including those conducted by \citet{kumwilaisak2008factors} and \citet{kurtz2011how}, have made strides in identifying the key predictive features for ICU LOS among patients with neurological conditions. These works contribute to the knowledge on feature selection and model performance in this domain. However, there is still a lack of a predictive model specifically designed to neurological ICU patients, which highlights the necessity of further research in this area.  

The MIMIC-IV dataset renders itself as a particularly powerful resource for addressing this research gap. Due to its large-scale, structured, and recentness, it provides an ideal foundation for developing predictive models tailored to ICU stays. MIMIC-IV includes rich clinical data, covering a wide range of neurological cases and enabling researchers to explore multiple use cases, relevant to this study’s objectives. By leveraging the strengths of this dataset, this research aims to develop a more accurate and clinically relevant predictive model for ICU LOS in patients with neurological disorders, thereby contributing to the advancement of patient care and hospital resource management.

\chapter{Methods}

\section{Data Preparation}

\subsection{Data Extraction from MIMIC-IV Dataset}

As previously mentioned, the MIMIC-IV dataset was chosen as the foundation for this study due to its extensive, high-quality, and the most recent (admissions from 2020 to 2022) data related to ICU patients. This dataset includes multiple interconnected tables, each containing valuable clinical, demographic, and procedural information. Given the vast amount of data available in the dataset, careful selection was necessary to ensure that only the most relevant tables were utilized for analysis.

For this research, a total of ten tables were identified and selected based on their significance in predicting the LOS for ICU patients. These tables provide crucial details such as patient demographics, vital signs, laboratory results, medical procedures, prescribed medications, and hospital admission records. By focusing on specified tables, the study aims to extract meaningful patterns while maintaining a streamlined and efficient data processing workflow.

\begin{figure}[H]
    \centering
    \includegraphics[width=0.8\linewidth]{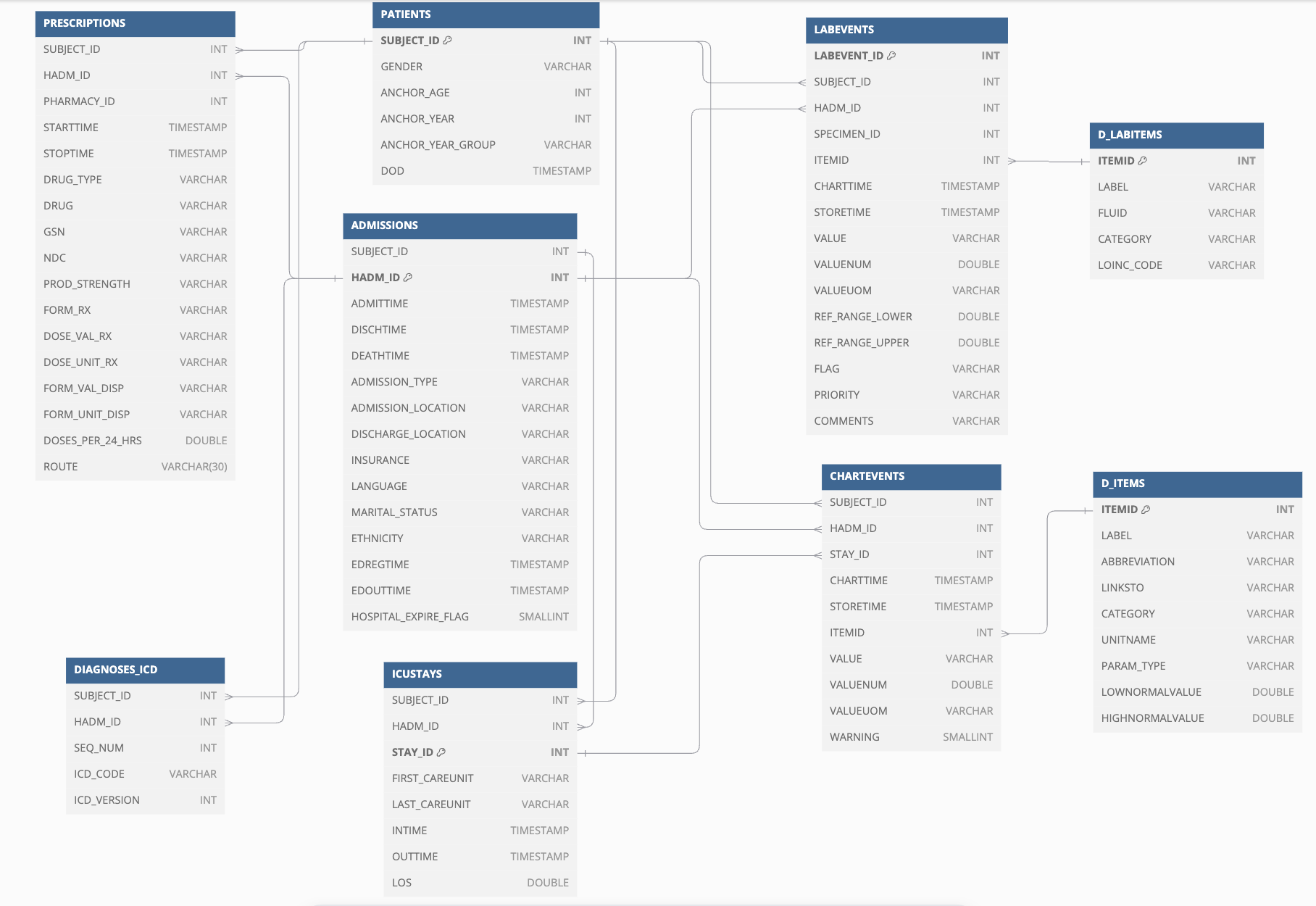}
    \caption{MIMIC-IV Tables for the Research}
    \label{fig:mimic_schema}
\end{figure}

The figure~\ref{fig:mimic_schema} represents the relationships between the selected tables. It shows the structure of each table and the established connections between these tables. 

\textit{Admissions} table serves as the central entity within the dataset, encapsulating comprehensive information related to each patient's hospitalization. This includes critical details such as \textit{insurance} type, \textit{language} spoken, \textit{marital\_status}, and \textit{race}, which may all contribute to broader social-demographic patterns influencing patient outcomes. Moreover, this table stores essential timestamps, including \textit{admission time}, \textit{discharge time}, and, when applicable, \textit{death time}, allowing for a complete temporal mapping of a patient’s hospital stay. Given its fundamental role, this table is further supported and enriched by several other key tables, including \textit{patients}, \textit{icustays}, and \textit{diagnoses\_icd}.  

The \textit{Patients} table contains personal and demographic data for each individual recorded in the database. Fields such as \textit{gender} and encoded \textit{age} (for patient anonymity) are included alongside metadata like \textit{year of admission} and \textit{year group}. Additionally, this table holds records of \textit{date of death}, which could also be useful for mortality analysis and understanding long-term outcomes of ICU patients.  

Among all tables, the \textit{Icustays} table plays the most significant role in this study, as it directly pertains to patient stays within the intensive care unit. It provides key attributes such as \textit{first} and \textit{last careunit}, which represent the different ICU departments a patient may have been transferred between. More importantly, it includes vital timestamps like \textit{intime} (ICU admission time) and \textit{outtime} (ICU discharge time), as well as the computed \textit{length of stay}. The primary objective of this research is to predict this length of stay accurately, making the \textit{icustays} table crucial to the study.  

The \textit{Diagnoses\_icd} table contains structured diagnostic data recorded during each hospital admission. It assigns International Classification of Diseases (ICD) codes to each patient’s medical condition, providing a standardized and consistent labeling system for various illnesses. These ICD codes are particularly useful during the data preparation phase, as they allow researchers to filter and identify patients based on specific diagnoses. In this study, this table is essential for selecting patients with neurological disorders, ensuring that the analysis remains focused on the targeted population.  

In addition to the diagnostic data, the \textit{prescriptions} table captures information about all medications prescribed during a patient's hospital stay. This data is valuable for evaluating possible correlations between drug administration and the duration of ICU stays. Certain medications might be indicative of more severe conditions requiring prolonged hospitalization, making this table an important component of predictive modeling.  

The dataset also includes two crucial time-series tables: \textit{labevents} and \textit{chartevents}. These tables document the results of various medical tests and bedside monitoring data collected throughout a patient’s ICU stay. Since many physiological variables fluctuate over time, these tables provide dynamic insights into a patient’s condition, which can be instrumental in predicting length of stay. Moreover, these tables also provide an opportunity to predict the remaining LOS based on the previous iterations/tests, which gives this study a unique chance to assess this approach.

To support these time-series data tables, two additional reference tables are included: \textit{d\_labevents} and \textit{d\_items}. The \textit{d\_labevents} table extends the \textit{labevents} table by providing labels for recorded test results, ensuring clarity in the interpretation of laboratory data. Similarly, the \textit{d\_items} table enhances the \textit{chartevents} table by providing descriptions for documented medical measurements, as well as lower and upper bounds for normal test values.

\subsection{Data Storage}

To effectively store and manage the extracted data for this study, the PostgreSQL database management system has been selected. PostgreSQL is an open-source relational database system that has been widely adopted in both academic research and industry applications. It is designed to provide efficient storage, retrieval, and management of structured data while ensuring reliability, scalability, and accordance with ACID (Atomicity, Consistency, Isolation, Durability) principles, making it a suitable choice for handling complex datasets like MIMIC-IV.  

One of the key advantages of PostgreSQL is its open-source nature, meaning that it is freely accessible to researchers and developers without the need for expensive licensing fees. This is particularly beneficial in an academic setting, where budget constraints may limit the availability of proprietary database solutions. Moreover, PostgreSQL offers sound support for SQL queries, indexing, and advanced data types, making it highly versatile for managing large-scale healthcare data \cite{postresql}.

For this particular study, the primary focus is on data extraction, integration, and analysis rather than optimizing query performance for real-time analytics. Since the study does not require extremely fast retrieval speeds or sophisticated parallel processing capabilities, there is no compelling need to deploy a high-end Online Analytical Processing (OLAP) system. OLAP systems are specifically designed for handling complex analytical queries, supporting large-scale data warehousing, providing high performance join-operations, and enabling vertical data storage. While they have the advantages in environments where frequent aggregations and multi-dimensional analysis are needed, their benefits do not significantly impact this study.  

Instead, PostgreSQL offers sufficient functionality for this study’s objectives. The ability to perform standard SQL-based queries ensures that data retrieval and transformation processes remain efficient.

Given that the main goal of this study is to extract and analyze structured patient data rather than implement an enterprise-level business intelligence solution, PostgreSQL emerges as the most practical choice.

After successfully selecting and deploying the PostgreSQL database system, the next crucial step involves designing an appropriate data storage model and determining the most effective approach for processing the data based on this structure. 

The first key consideration in this process is the nature of the experiments that will be conducted. Given that multiple types of experiments will be performed (some focusing on static data, others on time-series data, and some integrating static data with the time-series) it is essential to adopt a flexible storage strategy. To achieve this, the raw extracted data should be stored in its original format, ensuring that it remains readily accessible for future modifications. This approach would allow us to revisit the raw dataset and rebuild data marts if a new experiment requires significant changes in the way data is structured or processed. 

After that, the data mart layer is significant, so that data wouldn't need to be processed on each rerun of the experiment, but rather directly retrieved as a table, stored in data mart layer.

Once the raw data has been securely stored, the next critical layer is the data mart. Instead of reprocessing large volumes of data for each new experiment, the data mart enables the system to retrieve pre-processed and structured data efficiently. This means that once an experiment's dataset has been defined, cleaned, and structured, it can be stored in the data mart, significantly reducing computational overhead and improving workflow efficiency. 

\begin{figure}[H]
    \centering
    \includegraphics[width=0.8\linewidth]{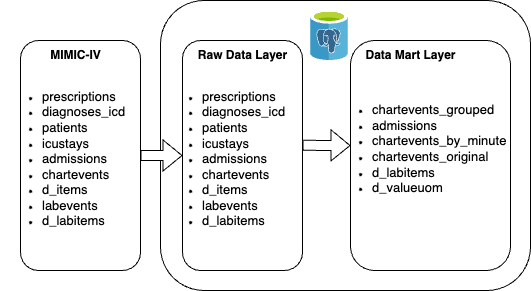}
    \caption{Data Storage Schema}
    \label{fig:data_storage}
\end{figure}

Figure~\ref{fig:data_storage} illustrates the structure of the data storage, outlining how information is organized and processed within the system. As discussed earlier, the Raw Data Layer serves as the foundation of the storage system, containing exact replicas of the nine afore mentioned key tables selected for the study. However, one specific modification is applied at this stage: a filtering mechanism is implemented to ensure that only data relevant to patients diagnosed with neurological disorders is retained.

The filtering process operates based on the ICD codes assigned to patient diagnoses for admissions. Specifically, an admission record is considered relevant if its primary diagnosis falls within one of the following ICD code patterns:

\begin{itemize}
    \item "I61\%" – Representing various types of intracerebral hemorrhage
    \item "I63\%" – Covering different forms of ischemic stroke
    \item "G41\%" – Corresponding to cases of status epilepticus
\end{itemize}

The percentage symbol ("\%") in these patterns acts as a wildcard, allowing for the inclusion of all subcategories within each diagnostic group. For instance, a diagnosis labeled "I619" (which denotes an \textit{unspecified intracerebral hemorrhage}) would be successfully captured by this filter. 

Next step is to create data marts. Before actually planning any transformations for that layer, our main goals needs to be addressed. 

Most of the related researches worked upon the data on the admission day (withing 24 hours after the admission). In order to assess the results of the current study, this approach must be tried, meaning the data from the \textit{admission}, \textit{icustays}, \textit{diagnoses\_icd}, and \textit{patients} tables should be combined as static data and then used to evaluate classic machine learning models, which have demonstrated strong performance in the related studies, namely: \textbf{Random Forest, SVM, XGBoost, and KNN}.  

Moreover, the recent advancements in transformer models, particularly \textbf{BERT} and \textbf{Temporal Fusion Transformer}, have significantly improved the ability to handle time-series data, opening up an opportunity to leverage additional information from \textit{chartevents} and \textit{labevents} tables.  

To achieve this, several scenarios can be considered. The two primary approaches include:  
\begin{itemize}  
    \item \textbf{Creating a table with a granularity of one minute}, where all relevant tests are placed as features (or columns in the table).  
    \item \textbf{Aggregating all test results} so that each row in the table represents a specific moment in time when at least one test was conducted on at least one patient.  
\end{itemize}  

Before actually building these two data marts, one additional step needs to be done: decide which tests (done in ICU) are relevant to the study. It is of high importance to do, as there are 3972 different tests, according to the \textit{d\_items} table, meaning that the high percentage of data as-is would not be used in the research. 

With the help of an expert in neurology and Chat-GPT, following groups of tests are identified as relevant.

\setlength{\LTpost}{-25pt}
\captionsetup[longtable]{skip=0pt}
\begin{center}
    \small
    \begin{longtable}{|p{3cm}|p{5cm}|p{5cm}|}
        \hline
        \textbf{Group Name} & \textbf{Tests} & \textbf{Relevance} \\
        \hline
        \endfirsthead

        \hline
        \textbf{Group Name} & \textbf{Tests} & \textbf{Relevance} \\
        \hline
        \endhead

        \hline
        \multicolumn{3}{r}{\textit{Continued on next page...}} \\  
        \hline
        \endfoot

        \hline
        \endlastfoot

        \textbf{Vital Signs and Hemodynamics} & HR, RR, ABPm, ABPd, ABPs, NBPm, NBPd, NBPs, Temperature C/F, Cerebral T (C), SpO2 Desat Limit &  
        Indicate overall patient stability, cardiovascular function, and risk of deterioration. Blood pressure and oxygenation are critical in stroke and brain injury patients. \\  
        
        \hline
        \textbf{Neurological Function and Consciousness Assessment} &  
        Eye Opening, Motor Response, Verbal Response, Pupil Response L/R, Motor Deficit, Delirium Assessment, CAM-ICU (Altered LOC, Disorganized Thinking, Inattention), Goal Richmond-RAS Scale, Daily Wake Up, Daily Wake Up Deferred &  
        Indicate brain injury severity and recovery trajectory. Prolonged impaired consciousness or delirium correlates with increased ICU LOS. \\  
        
        \hline
        \textbf{Blood Gas and Electrolyte Balance} &  
        PH (Arterial/Venous), HCO3 (Serum), Sodium, Potassium, Chloride, BUN &  
        Neurological patients often suffer from electrolyte imbalances affecting brain function, leading to complications such as cerebral edema and metabolic encephalopathy. \\  
        
        \hline
        \textbf{Ventilation and Respiratory Support} &  
        Ventilator Mode, Ventilator Mode (Hamilton), Ventilator Type &  
        Many neurological patients require mechanical ventilation due to coma or respiratory failure. Duration of ventilation significantly impacts ICU LOS. \\  
        
        \hline
        \textbf{Pain Management and Sedation} &  
        BIS - EMG, BIS Index Range, CPOT-Pain Assessment, CPOT-Pain Management, Pain Level, Pain Level Acceptable (Pre/Post Intervention), Pain Level Response, Pain Management, NMB Medication &  
        Pain and sedation management affect recovery speed. Over-sedation prolongs ICU stay, while inadequate pain control increases stress responses. \\  
        
        \hline
        \textbf{Patient-Controlled Analgesia (PCA) and Epidural Infusion} &  
        PCA (1-hour limit, attempt, basal rate, bolus, cleared, concentrations, dose, inject, lockout, medication, total dose), Epidural Infusion Rate, Epidural Medication &  
        Effective pain management reduces agitation and complications, impacting ICU LOS. Poor pain control can delay rehabilitation. \\  
        
        \hline
        \textbf{Neuromuscular Monitoring} &  
        TOF Response, TOF Twitch, Current Used/mA &  
        Used to monitor neuromuscular blockers, preventing prolonged paralysis and ensuring timely recovery. Essential for ventilated or post-seizure patients. \\  
        
        \hline
        \textbf{Liver and Coagulation Function} &  
        PTT, INR, AST, ALT, Direct Bilirubin, Total Bilirubin &  
        Liver dysfunction or coagulopathy can lead to increased ICU stay due to bleeding risks, drug metabolism issues, and hepatic encephalopathy. \\  
        
        \hline
        \textbf{Physical Therapy and Mobility} &  
        PT Splint Location \#1/\#2, PT Splint Status \#1/\#2 &  
        Early mobilization is key to reducing ICU-acquired weakness. Splinting is relevant in stroke or neuromuscular disorder patients. \\  

        \hline
    \end{longtable}
\end{center}

\captionof{table}{ICU Tests and Their Relevance to LOS Prediction for Neurological Patients}
\label{tab:ICU_tests}
\vspace{15pt}

After filtering by these tests (based on the \textit{d\_items} table's \textit{abbreviation} column), the result is joined with the \textit{chartevents} table, which stores detailed event-level data for each patient. This join operation ensures that only the relevant events corresponding to the specific tests are included. The data is then aggregated by \textit{hadm\_id} (representing the identifier of the admissions) and \textit{charttime} (representing the exact time when the test was taken). This aggregation process is essential to maintain a structured and time-sensitive view of the tests for each patient throughout their stay in the ICU. 

To address the tests, discussed above, two features for each of them will be added: \textbf{value of test} (when applicable) and \textbf{flag on whether the test results are within a norm range}. The value of the test represents the actual numerical measurement recorded, while the flag is a binary indicator showing whether the result falls within a predefined normal range for that specific test. This serves as a strong indicator of the patient's condition in real-time and might help models identify any potential abnormalities that may require clinical intervention resulting in the prolonged LOS.

Upon this aggregation, two marts are built: 

\begin{itemize}
    \item \textit{chartevents\_by\_minute}, which contains a row for each minute during each patient's stay in the ICU (regardless of whether tests were performed at that minute, meaning that there might be rows without any values for tests at certain times).
    \item \textit{chartevents\_original}, which has a row for each test the patient underwent (so, there is at least one result of the test present in each row, ensuring that all relevant lab test data is captured).
\end{itemize}

To support these two marts along with the \textit{admissions} mart, two dictionary tables are introduced: \textit{d\_labitems} and \textit{d\_valueoum}. The \textit{d\_labitems} table adds additional information on the tests, providing the type of fluid (e.g., joint fluid, pleural fluid, bone marrow, ascites, stool, cerebrospinal fluid, blood, urine, or other body fluids) and categorizing the tests based on the type of analysis performed. The \textit{d\_valueoum} table, on the other hand, provides the unit of measurement for each test result from the \textit{labevents} table, such as milligrams per deciliter (mg/dL) for glucose or millimoles per liter (mmol/L) for electrolytes, and knowing the unit of measurement is necessary for accurate interpretation of the test results. This also provides additional help in future data transformations which might take into account normalizing these parameters.

Together, these tables provide a comprehensive and structured way to analyze and interpret ICU test data, supporting the approaches mentioned previously. 

\section{Algorithms} \label{chap:algorithms}

As one of the results of the literature review, following algorithms were selected for this study, as they tend to achieve better results in the related work.

\begin{itemize}
    \item K-Nearest Neighbors
    \item Random Forest
    \item Support Vector Machine
    \item Gradient Boosting (XGBoost in particular)
\end{itemize}

\textbf{K-Nearest Neighbors} (or, simply, KNN) is an ML algorithm, which can be used both for the classification and the regression tasks. KNN takes \textit{number of neighbors} as the only hyperparameter, by which the label for the current object is decided. It works like that: for each of the \textit{k} neighbors the weight of 1/\textit{k} is assigned, for others it is zero. So that only these nearest neighbors are used in the calculation of the label for the input object. 

In general, algorithm does not consist of a \textit{training} step as it just saves the train data. When it is called to predict a label for a new object, it just calculates the distance to all other objects, select \textit{k} nearest neighbors and, if it a classification task, applies the "majority vote", otherwise return an average value of these \textit{k} nearest neighbors.

\textbf{K-Nearest Neighbors} (or, simply, KNN) is a simple ML algorithm that can be used for both \textbf{classification} and \textbf{regression} tasks. Unlike many machine learning algorithms, KNN is a \textit{lazy learning} method, meaning that it does not explicitly learn a model during training but instead in fact just stores all training instances and makes predictions based on them. 

KNN operates based on a single hyperparameter: \textbf{the number of neighbors} (\textit{k}), which determines how many closest data points influence the classification or regression decision. The algorithm follows these steps:

\begin{enumerate}
    \item \textbf{Store the training data:} all training samples are saved.
    \item \textbf{Compute distances:} when a new input object is given, the algorithm calculates the distance between this input and all objects provided in the training dataset. Default metric is Euclidean Distance:
        \[
        d(x, y) = \sqrt{\sum_{i=1}^{n} (x_i - y_i)^2}
        \]
        
    \item \textbf{Find the k nearest neighbors:} Select the \textit{k} closest data points based on the computed distances.
    \item \textbf{Predict the output:} 
    \begin{itemize}
        \item \textbf{For classification:} The label of the new input is calculated using a \textbf{majority vote} among the \textit{k} nearest neighbors.
        \item \textbf{For regression:} The predicted value is calculated as the \textbf{average} (or weighted average) of the values of the \textit{k} nearest neighbors.
    \end{itemize}
\end{enumerate}

By default, each of the \textit{k} neighbors is assigned an equal weight of \(1/k\), meaning they all contribute equally to the final decision. However, a more sophisticated approach assigns weights based on proximity, so that closer neighbors have more influence. One common method is using an inverse distance weighting:

\[
w_i = \frac{1}{d(x, x_i)}
\]

where \( d(x, x_i) \) is the distance between the input point and its neighbor \( x_i \). 

The choice of \textit{k} is crucial:
\begin{itemize}
    \item \textbf{Smaller k:} model is extremely sensitive to the noise and prune to overfitting.
    \item \textbf{Larger k:} model is smoother, but quite likely to underfit.
\end{itemize}
A common approach is to use cross-validation to find the optimal value of \textit{k}. \cite{fix1989knn}
\vspace{15pt}

Another algorithm, \textbf{Random Forest}, is built upon the Decision Tree, which is a supervised ML algorithm. A Decision Tree consists of three main components:  
\begin{itemize}
    \item Root Node – The starting root of the created tree, which represents the dataset.
    \item Branches – The decision paths to split data from the dataset by certain conditions.
    \item Leaf Nodes – The terminal nodes which provide the predicted output.
\end{itemize}

Distribution of the training data between branches is handled by selecting the most relevant splitting criteria. The goal is to minimize either the Gini impurity or entropy, ensuring that the data is divided in a way that maximizes accuracy. For example, Gini impurity is calculated as:

\[
G = 1 - \sum_{i=1}^{C} p_i^2
\]

where \( p_i \) is the probability of class \( i \) and \( C \) is the total number of classes.

Entropy is based on the concept of information gain from information theory:

\[
H = -\sum_{i=1}^{C} p_i \log_2 p_i
\]

where \( p_i \) is the probability of class \( i \). Lower entropy  means purer nodes.

In general, Random Forest algorithm is an ensemble learning method that builds multiple Decision Trees. Instead of relying on a single tree’s prediction, Random Forest aggregates the outputs of multiple trees, where the final result is determined through a voting mechanism (for classification tasks) or an average prediction (for regression tasks). This reduces overfitting and enhances the model’s overall robustness and accuracy. 

So, the algorithm follows these steps:
\begin{enumerate}
    \item \textbf{Bootstrap Sampling:} The algorithm creates multiple Decision Trees using different random subsets of the training data (with replacement).
    \item \textbf{Feature Randomization:} Instead of considering all features for each split, only a random subset of features is chosen at each split. This reduces correlation between trees.
    \item \textbf{Tree Construction:} Each Decision Tree is grown independently using the subset of training data and features.
    \item \textbf{Aggregation:} The final prediction is calculated as average of the trees' predictions (for regression) or majority voting (for classification).
\end{enumerate}

Given a dataset \( D = \{ (x_1, y_1), (x_2, y_2), ..., (x_n, y_n) \} \), a Random Forest model consists of \( B \) individual decision trees, where each tree \( T_b \) is trained on a bootstrap sample \( D_b \).

\textbf{For classification:} The final predicted class \( \hat{y} \) is determined using majority voting:

\[
\hat{y} = \arg\max_k \sum_{b=1}^{B} \mathbb{1}(T_b(x) = k)
\]

where \( \mathbb{1}(\cdot) \) is an indicator function that counts the votes for class \( k \).

\textbf{For regression:} The final predicted value \( \hat{y} \) is computed as the average of predictions from all trees:

\[
\hat{y} = \frac{1}{B} \sum_{b=1}^{B} T_b(x)
\]

Naturally, \textit{number of estimators} (amount of Decision Trees to build for a Random Forest), \textit{max depth} (of each Decision Tree), \textit{minimum samples to split}, \textit{minimum samples in a leaf} are the hyperparameters, which optimization needs to be conducted before presenting the final Random Forest model in this study.
\cite{breiman2001rf}

\vspace{15pt}

\textbf{Support Vector Machine (SVM)} is a supervised learning algorithm used for both classification and regression tasks. It is mostly effective in high-dimensional spaces and is widely applied to complex pattern recognition problems.

SVM is based on the idea of finding the optimal \textit{hyperplane} that best separates the data into different classes. The aim of the algorithm is to maximize the distance between the separating hyperplane and the closest data points (so called \textit{support vectors}, hence the name of the algorithm). 

Given a labeled dataset:

\[
D = \{ (x_1, y_1), (x_2, y_2), ..., (x_n, y_n) \}, \quad y_i \in \{-1, +1\}
\]

where \( x_i \) represents feature vectors and \( y_i \) represents class labels, the goal of SVM is to find a hyperplane:

\[
w^T x + b = 0
\]

where:
\begin{itemize}
    \item \( w \) is the weight vector,
    \item \( x \) is the input feature vector,
    \item \( b \) is the bias term.
\end{itemize}

For linearly separable data, SVM finds the optimal hyperplane by maximizing the margin:

\[
\max_{w, b} \frac{2}{\|w\|}
\]

subject to the constraint:

\[
y_i (w^T x_i + b) \geq 1, \quad \forall i
\]

This ensures that all data points are correctly classified with the maximum margin.

When the data is not perfectly separable, a \textbf{slack variable} \( \xi_i \) is introduced to allow some object with falsely assigned classes:

\[
y_i (w^T x_i + b) \geq 1 - \xi_i, \quad \xi_i \geq 0
\]

Then the objective function then becomes:

\[
\min_{w, b} \frac{1}{2} \|w\|^2 + C \sum_{i=1}^{n} \xi_i
\]

where \( C \) acts as a regularization parameter that balances between maximizing the margin and allowing classification errors.

If the data is not linearly separable, SVM can use a \textit{kernel function} \( K(x_i, x_j) \) to transform the input space into a higher-dimensional feature space, making it easier to separate. 

While being a solid algorithm for the classification tasks, SVM can also be used for regression tasks, known as \textit{Support Vector Regression (SVR)}. SVR, unlike SVM, aimes at finding a function that approximates the target variable with minimal deviation.

The objective function for SVR is:

\[
\min_{w, b} \frac{1}{2} \|w\|^2 + C \sum_{i=1}^{n} (\xi_i + \xi_i^*)
\]

subject to:

\[
|y_i - (w^T x_i + b)| \leq \epsilon + \xi_i, \quad \xi_i, \xi_i^* \geq 0
\]

where \( \epsilon \) is a margin of tolerance for errors. \cite{cortes1995support}
\vspace{15pt}

The last of the considered ML algortihms is an implementation of \textbf{Gradient Boosting}, namely XGBoost. Gradient Boosting emerges as a powerful ensemble learning algorithm that builds models sequentially, so it tries to minimize the errors of the previously built models, iteratively improving the quality. Two popular implementations of Gradient Boosting are \textbf{XGBoost} and \textbf{CatBoost}, Although the latter one was not mentioned in the related works, it is still considered in this study because of the demonstrated state-of-the-art performance in the real-world applications. 

In general, Gradient Boosting is an ensemble method that builds a model in a stage-wise fashion by optimizing a loss function. It sequentially fits weak learners (typically Decision Trees with relatively small depth) to minimize loss during training from the previous iteration.

Given a dataset:

\[
D = \{ (x_1, y_1), (x_2, y_2), ..., (x_n, y_n) \}
\]

where \( x_i \) represents feature vectors and \( y_i \) represents target values, the goal of Gradient Boosting is to model the function \( F(x) \) that minimizes the loss \( L(y, F(x)) \).

The Gradient Boosting algorithm follows these steps:

\begin{enumerate}
    \item Initialize the model with a constant value:
    \[
    F_0(x) = \arg\min_c \sum_{i=1}^{n} L(y_i, c)
    \]
    \item For \( m = 1 \) to \( M \) (number of boosting rounds):
    \begin{itemize}
        \item Compute the negative gradient:
        \[
        r_{im} = - \frac{\partial L(y_i, F(x_i))}{\partial F(x_i)}
        \]
        \item Fit a Decision Tree to approximate \( r_{im} \), giving a new model \( h_m(x) \).
        \[
        h_m(x) = \arg\min_c \sum_{i=1}^{n} L(r_{im}, c)
        \]
        \item Update the model:
        \[
        F_m(x) = F_{m-1}(x) + \eta h_m(x)
        \]
        where \( \eta \) is the learning rate.
    \end{itemize}
    \item The final model is:
    \[
    F_M(x) = F_0(x) + \sum_{m=1}^{M} \eta h_m(x)
    \]
\end{enumerate}

\textbf{XGBoost (Extreme Gradient Boosting)} is an optimized implementation of Gradient Boosting that improves training efficiency and model performance \cite{chen2016xgboost}. It introduces several enhancements:

\begin{itemize}
    \item \textbf{Regularization:} XGBoost minimizes the following objective:
    \[
    \sum_{i=1}^{n} L(y_i, \hat{y}_i) + \sum_{t=1}^{T} \left( \gamma T + \frac{1}{2} \lambda \sum_{j} w_j^2 \right)
    \]
    where \( T \) is the number of leaves, \( \gamma \) controls tree complexity, and \( \lambda \) penalizes large weights to reduce overfitting.
    
    \item \textbf{Second-Order Approximation:} Instead of using only the gradient, XGBoost uses the second-order Taylor expansion of the loss function:
    \[
    L(y, F) \approx L(y, F_{m-1}) + g h_m + \frac{1}{2} h h_m^2
    \]
    where \( g \) and \( h \) are the first and second derivatives of the loss function.
    
    \item \textbf{Handling Missing Values:} XGBoost can automatically learn optimal split directions for missing data.
\end{itemize}

CatBoost is another Gradient Boosting implementation specifically designed to handle categorical features efficiently \cite{prokhorenkova2018catboost}. It introduces:

\begin{itemize}
    \item \textbf{Ordered Boosting:} Traditional Gradient Boosting suffers from overfitting when categorical features are encoded. CatBoost avoids this by using ordered target statistics.
    \item \textbf{Efficient Categorical Encoding:} CatBoost transforms categorical features into numerical representations using a technique based on conditional probability.
    \item \textbf{Symmetric Trees:} Unlike standard decision trees, CatBoost grows balanced trees, reducing overfitting and improving training speed.
\end{itemize} 

\newpage
\begin{center}
    \begin{tabular}{|c|c|c|}
        \hline
        \textbf{Feature} & \textbf{XGBoost} & \textbf{CatBoost} \\
        \hline
        \textbf{Categorical Handling} & Not implemented & Ordered Target Statistics \\
        \hline
        \textbf{Tree Structure} & Unrestricted & Symmetric Trees \\
        \hline
    \end{tabular}
\end{center}
\captionof{table}{Comparison of XGBoost and CatBoost}

\vspace{15pt}

To achieve optimal performance, tuning hyperparameters is essential. Since XGBoost is used in this study based on Decision Trees, it shares several hyperparameters with Random Forest, such as the \textit{number of estimators}, \textit{maximum tree depth}, and \textit{subsample size}.  

Additionally, XGBoost introduces other parameters, including the \textit{learning rate}, which controls the step size during training (a higher value speeds up learning but increases the risk of missing the optimal solution). Other important parameters include \textit{column sample by tree} (determining the fraction of features used per tree), \textit{gamma} (minimum loss reduction required for a split), \textit{reg alpha} (L1 regularization for feature selection), and \textit{reg lambda} (L2 regularization for controlling model complexity).

Because of the similarity between XGBoost and CatBoost, latter shares most of the hyperparameters of former. However, there are 2 more, that are going to be considered in this study: \textit{number of splits}, named as border\_count, and \textit{randomness in bagging}. \textit{Number of splits} defines the number of bins (also called borders) for numerical feature quantization, which affects how continuous features are transformed into categorical-like features. \textit{Bagging temperature} acts as a level control of randomness in boostrapping, as CatBoost applies Bayesian bootstrapping under the hood. That means, that the probability of selecting a sample is proportional to a gamma distribution:

\[
P(x_i) \sim \text{Gamma}(\alpha, 1), \quad \alpha = \frac{1}{\texttt{bagging\_temperature}}
\]

Lowering this parameter closer to zero results in deterministic sampling (leads to less variance, but highers the risk of overfitting), while higher values close to 1 result in a uniform sampling (leads to a higher variance, but prevents overfitting).

\vspace{15pt}
Given that the MIMIC-IV dataset consists of rich time-series data (\textit{chartevets} table in particular), it presents an opportunity to apply advanced ML techniques, particularly \textbf{Long Short-Term Memory (LSTM)} models. By leveraging the sequential data, LSTM models can capture the temporal dependencies and patterns underlying in patient data over time, allowing for more accurate predictions and assessments, which could also be helpful in medical settings leading to possibly reliable results in predicting LOS of ICU stays.

Generally, the idea of recurrently processing data is implemented in Recurrent Neural Networks (RNNs). RNNs are a class of neural networks which is designed to process sequential data by maintaining a hidden state that captures information from previous steps. This ability makes them suitable for tasks like time-series analysis. However, the most consistent results of RNN-based models were achieved by LSTMs \cite{yu2019lstm}.

Originally, LSTMs were developed to address limitations of standard RNNs, such as \textit{long-term dependencies}. Key feature of the proposed class of new models is LSTM cell. Such cell consists of several components:

\begin{itemize}
    \item \textbf{Cell State} ($C_t$): Acts as a memory buffer, allowing information to flow unchanged across time steps.
    \item \textbf{Hidden State} ($h_t$): Represents the output of the LSTM cell at each time step.
    \item \textbf{Gates}: Mechanisms that regulate the flow of information:
    \begin{itemize}
        \item \textbf{Forget Gate}: Decides which information to discard from the cell state:
        \[
        f_t = \sigma(W_f \cdot [h_{t-1}, x_t] + b_f)
        \]
        \item \textbf{Input Gate}: Determines what new information to add to the cell state:
        \[
        i_t = \sigma(W_i \cdot [h_{t-1}, x_t] + b_i)
        \]
        \[
        \tilde{C}_t = \tanh(W_C \cdot [h_{t-1}, x_t] + b_C)
        \]
        \item \textbf{Output Gate}: Computes the output of the LSTM cell:
        \[
        o_t = \sigma(W_o \cdot [h_{t-1}, x_t] + b_o)
        \]
    \end{itemize}
\end{itemize}

The cell state is updated as:
\[
C_t = f_t \ast C_{t-1} + i_t \ast \tilde{C}_t
\]
The hidden state is computed as:
\[
h_t = o_t \ast \tanh(C_t)
\]
where $\sigma$ is the sigmoid activation function and $\ast$ represents element-wise multiplication.

LSTMs proved to be efficient in the medical setup \cite{sun2020irrts}, specifically on the MIMIC-III dataset irregular time series data, therefore this algorithm will be assessed in the current study.

\vspace{15pt}
Regarding the algorithms, which can work with time-series data, as previously discussed, two were selected: BERT and Temporal Fusion Transformer.

\textbf{BERT (Bidirectional Encoder Representations from Transformers)} is a deep learning model which is based on the \textbf{Transformer} architecture, introduced by  \citet{vaswani2017attention}. Unlike sequential models such as RNNs and LSTMs, which process input step-by-step, BERT processes input sequences in parallel using \textit{self-attention}, making it highly efficient at capturing long-range dependencies.

Key Components of BERT:
\begin{itemize}
    \item \textbf{Token Embeddings} – Converts input tokens into dense vector representations.
    \item \textbf{Positional Encoding} – Adds order information to input sequences since Transformers do not have implemented sequential awareness.
    \item \textbf{Multi-Head Self-Attention} – Learns different relations between all time steps simultaneously. 
    \begin{figure}[H]
        \centering
        \includegraphics[width=0.5\linewidth]{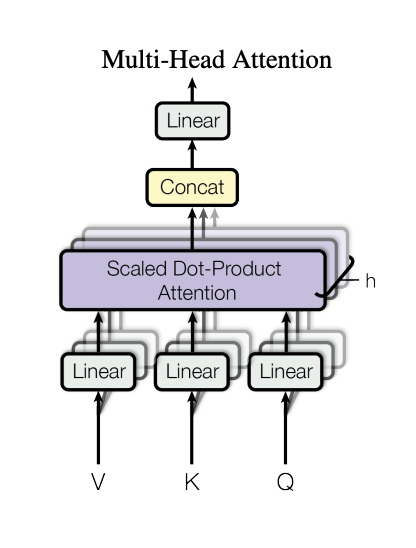}
        \caption{Multi-Head Attention}
        \label{fig:enter-label}
    \end{figure}
    \item \textbf{Feedforward Layers} – Applies non-linear transformations after attention layers.
    \item \textbf{Hidden Layers} – Stacked  blocks that iteratively update and enhance the learned representations.
    \item \textbf{Layer Normalization \& Residual Connections} – Stabilizes training and enables deep stacking of layers.
\end{itemize}

The \textit{self-attention} mechanism allows BERT to model dependencies between all elements in a sequence efficiently. Given an input sequence \( X \in \mathbb{R}^{n \times d} \), where \( n \) is the sequence length and \( d \) is the embedding dimension, self-attention is computed as follows:

\begin{enumerate}
    \item \textbf{Compute Query, Key, and Value Matrices}  
    Each input vector \( x_i \) is projected into three different spaces:
    \[
    Q = XW^Q, \quad K = XW^K, \quad V = XW^V
    \]
    where \( W^Q, W^K, W^V \in \mathbb{R}^{d \times d_k} \) are learnable weight matrices for queries, keys, and values.

    \item \textbf{Compute Attention Scores}  
    The attention scores are computed using scaled dot-product attention:
    \[
    \text{Attention}(Q, K, V) = \text{softmax} \left( \frac{QK^T}{\sqrt{d_k}} \right) V
    \]
    where the scaling factor \( \sqrt{d_k} \) prevents large values that could lead to vanishing gradients.

    \item \textbf{Multi-Head Self-Attention}  
    Instead of computing a single attention mechanism, BERT applies \textbf{multi-head attention}, which runs multiple self-attention computations in parallel:
    \[
    \text{MultiHead}(Q, K, V) = \text{Concat}(\text{head}_1, ..., \text{head}_h) W^O
    \]
    where each head \( \text{head}_i \) corresponds to an independent self-attention computation, and \( W^O \) is a learned weight matrix.

    \item \textbf{Feedforward Layer}  
    Each attention output is passed through a position-wise feedforward network (FFN):
    \[
    \text{FFN}(x) = \max(0, xW_1 + b_1) W_2 + b_2
    \]
    This introduces non-linearity and enhances the model’s capacity to learn complex representations.
\end{enumerate}

Although BERT was originally designed for NLP, its \textit{self-attention mechanism} makes it highly suitable for time-series analysis. There are several reasons supporting that:

\begin{itemize}
    \item \textbf{Capturing Long-Term Dependencies}: BERT’s self-attention mechanism can \textit{capture dependencies between distant time points} without being constrained by sequential processing.

    \item \textbf{Parallelization and Computational Efficiency}: Unlike RNNs, which process data sequentially, BERT applies \textit{self-attention} to all time steps simultaneously. This enables faster training and inference, making it more efficient for large-scale time-series datasets.
\end{itemize}

Another recently developed by \citet{lim2021tft} transformer-based model, \textbf{Temporal Fusion Transformer (TFT)}, also emerges as one of the most promising algorithms to deal with the time-series data. This algorithm uses combination of attention-based  and recurrent mechanisms to handle short and long-term dependencies. At the same time this transformer is also well-equipped for the feature importance analysis.

Main idea is to master multi-horizon forecasting, 
with that aim TFT was developed to comprise of following components:

\begin{itemize}
    \item \textbf{Gating Mechanisms} which are used to selectively skip those components, which did not get enough usage. In other words, the assumption is - simpler models might be beneficial. Therefore, \textit{Gated Residual Network} was proposed:
    The Gated Residual Network (GRN) is defined as:

    \begin{equation}
        \text{GRN}_\omega (a, c) = \text{LayerNorm} \left( a + \text{GLU}_\omega(\eta_1) \right),
    \end{equation}
    where
    \begin{equation}
        \eta_1 = W_{1,\omega} \eta_2 + b_{1,\omega},
    \end{equation}
    \begin{equation}
        \eta_2 = \text{ELU} \left( W_{2,\omega} a + W_{3,\omega} c + b_{2,\omega} \right).
    \end{equation}
    
    ELU (Exponential Linear Unit) is an activation function, and $\eta_1, \eta_2 \in \mathbb{R}^{d_{\text{model}}}$ are intermediate layers. LayerNorm acts as a normalization layer method, while $\omega$ denotes weight sharing.

    Upon that, \textit{Gated Linear Units} are built in order to ignore any parts of architecture, that are not relevant for the current task.

    \item \textbf{Variable Selection Networks} which are used to identify the most relevant inputs for the static and time-depending variables. Variable selection weights are generated by the transformed input of variable and a context vector $c_s$ through a GRN. Then a Softmax layer is applied:

    \begin{equation}
        v_{\chi t} = \text{Softmax} (\text{GRN}_{v_{\chi}} (\Xi_t, c_s)),
    \end{equation}
    
    where $v_{\chi t} \in \mathbb{R}^{m_{\chi}}$ is a vector of variable selection weights, and $c_s$ is generated by a static covariate encoder.

    \item \textbf{Static Covariate Encoders} which are used to integrate the static variables into the Neural Network. This component is a key difference from other transformers as it allows to enrich time-series with a static data.

    \item \textbf{Temporal Processing} which is used to capture the short and long-term trends. So this component acts similar to the BERT mechanism and enables time-series processing by using multi-head attention.

    \item \textbf{Prediction Intervals} which are used to estimate the range of the predicted variable at each prediction time mark.

\end{itemize}

\section{Metrics}
Since the task is a classification problem, appropriate evaluation metrics will be selected. Moreover, to assure comparability with previous research (despite differences in specific objectives but within a similar domain) the chosen metrics align with those used in relevant studies.

Most commonly used are:

\begin{itemize}
    \item \textbf{Accuracy} – The overall correctness of the model by calculating the ratio of correctly predicted labels to the total number of objects in the testing data. Since it is stated earlier, that in this study classification with three classes (\(C_1, C_2, C_3\)) is employed, formula is adjusted to this case. Therefore, accuracy is defined as:  
    \[
    \text{Accuracy} = \frac{TP_1 + TP_2 + TP_3}{TP_1 + TP_2 + TP_3 + FP_1 + FP_2 + FP_3 + FN_1 + FN_2 + FN_3}
    \]
    where \( TP_i \) refers to the correctly predicted instances for class \( C_i \), while \( FP_i \) and \( FN_i \) are false positives and false negatives for each class, respectively.

    \item \textbf{Precision} – Represents the proportion of correctly predicted instances of a specific class among all instances predicted as that class. Precision for class \( C_i \) is given by:  
    \[
    \text{Precision}_i = \frac{TP_i}{TP_i + FP_i}
    \]

    \item \textbf{Recall} – The ability of the model to correctly identify instances of a given class from all actual instances of that class. Recall for class \( C_i \) is defined as:  
    \[
    \text{Recall}_i = \frac{TP_i}{TP_i + FN_i}
    \]

    \item \textbf{F1 Score} – A harmonic mean of precision and recall, computed per class and then averaged. The F1 score for class \( C_i \) is given by:  
    \[
    \text{F1 Score}_i = 2 \times \frac{\text{Precision}_i \times \text{Recall}_i}{\text{Precision}_i + \text{Recall}_i}
    \]
    The overall F1 Score is commonly aggregated using macro or micro averaging \cite{grandini2020metrics}:
    \[
    \text{F1 Score}_{\text{macro}} = \frac{1}{3} \sum_{i=1}^{3} \text{F1 Score}_i
    \]
    \[
    \text{F1 Score}_{\text{micro}} = \sum_{i=1}^{3} w_i \times \text{F1 Score}_i
    \]
    where \( w_i \) represents the proportion of instances belonging to class \( C_i \), meaning that the macro approach treats each class as the same, while micro addresses the sizes of classes. The latter helps understand the quality of model better in case of class imbalance.
\end{itemize}

Although there are two ways to produce the output as a single number (\textit{macro} and \textit{micro}, as stated before), the end results will be calculated as an average between 3 classes in order to ease comparability, while the outcomes for each different class will be discussed in the results chapter. It is a preferred approach since there is no significant class imbalance in the current study.

\section{Experiments}

Given that the study has several settings to evaluate models mentioned in the previous chapter, a pipeline is proposed. First, the data from the built data mart \textit{admissions} is used for the KNN, Random Forest, SVM, XGBoost and CatBoost models to assess their performance. Then the data from \textit{chartevents\_original} is used to assess LSTM model's ability to predict the remaining LOS based on the fixed window of events in the past. Likewise, \textit{chartevents} data will be used for examining capabilities of BERT-based model to calculate the remaining LOS. Finally, the static data of \textit{admissions} will be combined with the time series from \textit{chartevents} to train Temporal Fusion Transformer.

\subsection{Experiments on Static Data}

After preparing the \textit{admissions} data mart, afore mentioned classic ML models can be assessed to predict total LOS in ICU for patients with neurological disorders, based on their admission data and test results. As mentioned before, the task is switched to a classification one by applying the following transformation to the LOS column:

\begin{itemize}
    \item \textbf{Short}: 0–2 days
    \item \textbf{Medium}: 3–7 days
    \item \textbf{Long}: More than 7 days
\end{itemize}

Given the data on admissions, this classification ensures that there is no extreme class imbalance in place.

\begin{figure}[H]
    \centering
    \includegraphics[width=0.2\linewidth]{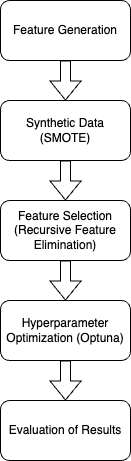}
    \caption{Proposed Pipeline}
    \label{fig:proposed_pipeline}
\end{figure}

Figure~\ref{fig:proposed_pipeline} above illustrates the sequence of steps to evaluate classic ML models.

\begin{enumerate}
    \item \textbf{Feature Generation}: \textit{Admissions} table is enhanced with the tests results along with the admission daytime (night, morning, day, evening) and admission month.
    \item \textbf{Synthetic Data (SMOTE)}: given that there is only 6941 admissions of patients with diagnosed neurological disorder, the data augmentation is aimed at providing more data for models to rise their possible generalization. SMOTE algorithm creates a combination of feature values between real minority classes instances and their neighbors \cite{chawla2002smote}. To successfully apply SMOTE algorithm, in this research missing values are handled in beforehand during the data preparation step. 
    SMOTE method works as follows: it identifies the minority class, selects a random sample from it, compute the nearest neighbors with a given metric and then creates new synthetic sample with the help of linear extrapolation:
    \begin{equation}
        x_{\text{new}} = x_{\text{minority}} + \lambda \cdot (x_{\text{neighbor}} - x_{\text{minority}})
    \end{equation}
    where \( \lambda \) is a random number in the range between 0 and 1, ensuring that the synthetic sample is on the line segment between the original sample and its neighbor.
    \item \textbf{Feature Selection}: after feature generation step, \textit{admissions} data contains 332 different features. Therefore dimensionality reduction is needed. Feature Selection is handled by applying \textit{Recusive Feature Elimination}. This algorithm trains a model on all the provided features and then sequentially eliminates \textit{k} features at each step, which result in higher model performance comparing with the previous array of features \cite{chen2007rfe}.
    \item \textbf{Hyperparameter Optimization}: renders as an important step, due to enormous number of combinations of hyperparameters that can be tested for a model. Therefore, to develop an optimal model this step needs to be done to select the best hyperparameter combination. In this study \textbf{Optuna} framework is employed for such task. Optuna applies advanced sampling and pruning techniques \cite{akiba2019optuna}, making it state of the art solution \cite{jafar2023comparative}.
    \item \textbf{Evaluation of Results}: is the last step, which critically assesses the outcomes of trained model.
\end{enumerate}

This pipeline will be applied to each model independently, except for the evaluation part, which will be done in comparison between all classic ML models. As discussed in the \textit{Algorithms} section, these experiments on hyperparameters will be conducted in order to find best possible combination of them:

\begin{itemize}
    \item \textbf{KNN}: 
        \begin{enumerate}
            \item \textbf{Number of neighbors}: range between 1 and 50.
            \item \textbf{Weight}: \textit{uniform} or \textit{distance}.
            \item \textbf{Metric}: \textit{Eucledian} or \textit{Manhattan}.
        \end{enumerate}
    \item \textbf{SVM}: 
        \begin{enumerate}
            \item \textbf{C}: range between 1e-3 and 1e3.
            \item \textbf{Kernel}: \textit{linear}, \textit{rbf}, \textit{poly}, or \textit{sigmoid}.
            \item \textbf{Gamma}: if kernel is \textit{linear}, then \textit{scale}, otherwise range between 1e-4 and 1e1.
            \item \textbf{Degree}: if kernel is \textit{poly}, then range between 2 and 5, otherwise 3.
        \end{enumerate}
    \item \textbf{Random Forest}: 
        \begin{enumerate}
            \item \textbf{Number of estimators}: range between 10 and 2000.
            \item \textbf{Maximum depth of trees}: range between 3 and 50.
            \item \textbf{Minimum samples to split}: range between 2 and 20.
            \item \textbf{Minimum samples in a leaf}: range between 1 and 10.
            \item \textbf{Maximum features to consider}: \textit{sqrt}, \textit{log2} or no restriction.
        \end{enumerate}
    \item \textbf{XGBoost}: 
        \begin{enumerate}
            \item \textbf{Number of estimators}: range between 10 and 2000.
            \item \textbf{Maximum depth of trees}: range between 3 and 50.
            \item \textbf{Learning rate}: range between 0.00001 and 0.3.
            \item \textbf{Subsample ratio of training instances}: range between 0.3 and 1.
            \item \textbf{Column subsample by tree}: range between 0.5 and 1.
            \item \textbf{Minimum loss reduction}: range between 0 and 5.
            \item \textbf{L1 regularization}: range between 0 and 20.
            \item \textbf{L2 regularization}: range between 0 and 20.
            \item \textbf{Minimum child weight}: range between 1 and 10.
        \end{enumerate}
        \item \textbf{CatBoost}: 
        \begin{enumerate}
            \item \textbf{Number of boosting iterations}: range between 10 and 2000.
            \item \textbf{Maximum depth of trees}: range between 3 and 50.
            \item \textbf{Learning rate}: range between 0.00001 and 0.3.
            \item \textbf{L2 regularization}: range between 1 and 10.
            \item \textbf{Number of splits for numerical features}: range between 32 and 255.
            \item \textbf{Subsample temperature}: range between 0 and 1.
            \item \textbf{Random noise applied to features}: range between 0 and 10.
        \end{enumerate}
\end{itemize}

\subsection{Experiments on Time Series Data}

Time series data is used for the LSTM and BERT-based approaches in this study.

Before starting directly applying the data to the model, several steps need to be implemented first:

\begin{itemize}
    \item \textbf{Handle Missing Data}: since the \textit{chartevents} marts are constructed as the aggregates with newly created columns for each test, these tables are quite sparse (rarely had a patient undergo more than 1 test at a time). Therefore this issue must be addressed. Since it a common issues for the researchers working with time series data, that problem has gained significant attention. One of the most common approaches, linear extrapolation, proved to be solid, however, its usefulness varies depending on the domain of data \cite{chow1971extrapolation}.
    \item \textbf{Data Augmentation}: given that there is information only on 6941 patients with neurological disorders, meaning that there is the same amount of patients' time series in \textit{chartevents} table, data needs to be enhanced. One common approach to avoid generating synthetic data and only rely on the real data, is to enhance data with sliding window. 
    \begin{figure}[H]
        \centering
        \includegraphics[width=0.8\linewidth]{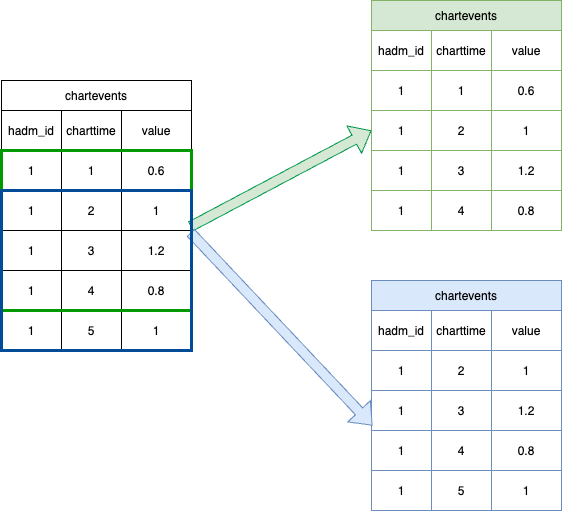}
        \caption{Sliding Window Demonstration}
        \label{fig:sliding_window}
    \end{figure}
    Figure~\ref{fig:sliding_window} illustrates the principle of sliding window. It has two parameters: \textit{length} (how many records show a window consist of) and \textit{step} (how much should the next window be shifted from the previous one). In that example, sliding window is defined with the length of 4 and the step of 1.
    
\end{itemize}

As for the considered sizes of \textit{window} and \textit{step} for the sliding window approach, firstly the power distribution of number of rows for each patient is analyzed. Based on that, list of window sizes and list of steps are proposed.
\begin{figure}[H]
    \centering
    \includegraphics[width=0.8\linewidth]{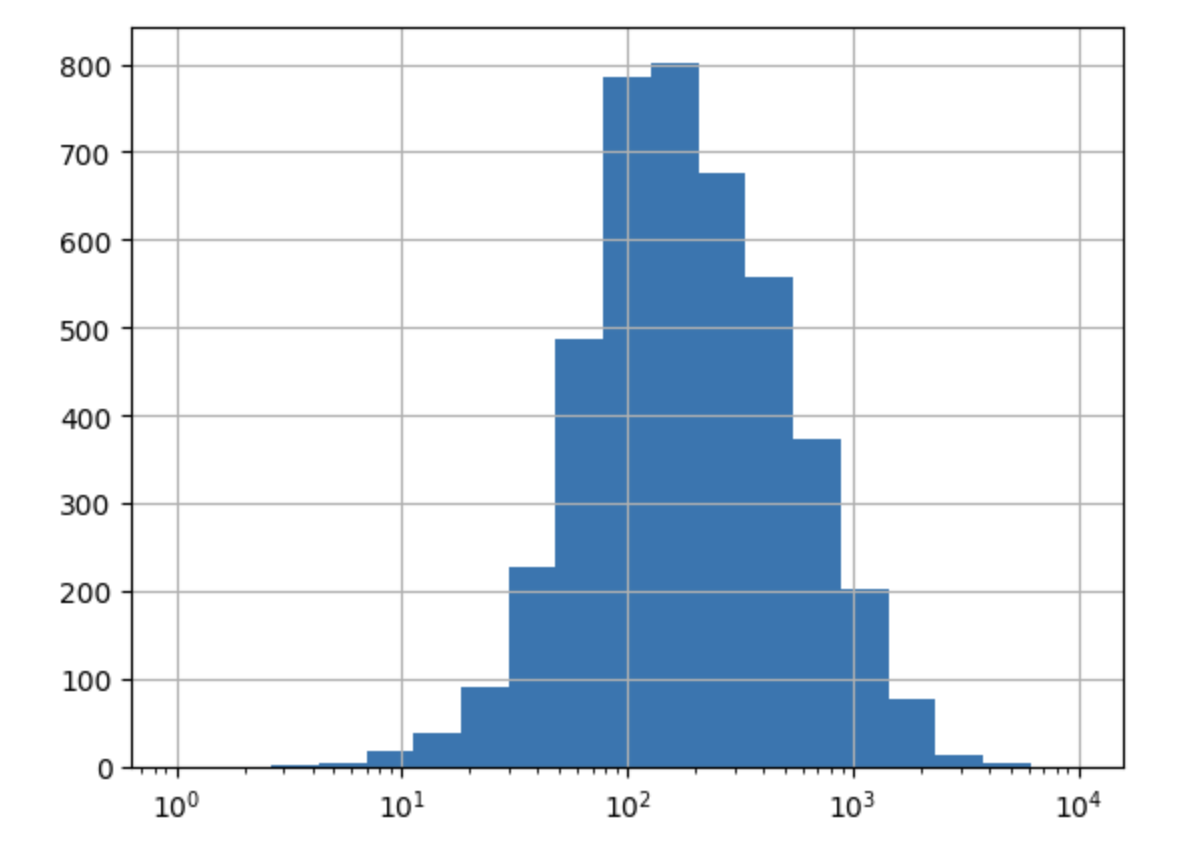}
    \caption{Power Distribution of Number of Patients Records in Chartevents Table}
    \label{fig:powerdistribution}
\end{figure}

According to the Figure~\ref{fig:powerdistribution} most of the patients do not have more than 1000 records, meaning that the upper bound of the window size should be 1000 (otherwise this approach does not produce additional data).

Regarding step sizes, there is more flexibility as there are no limitations. Therefore, this study will assess the step sizes by increasing them until the deterioration of metrics. 

Window and step sizes are often selected in a fashion so that they are represented by the numbers of power of 2, as it simplifies the implementation and reduces the number of calculations \cite{koc1995swtechniques}. Therefore, with the constraints mentioned above, following sizes will be considered:

\begin{itemize}
    \item \textbf{Window sizes}: 64, 128, 256, 512, 1024.
    \item \textbf{Step sizes}: 32, 64, 128, 256.
\end{itemize}

While this study will only assess the influence of window/step sizes on the LSTM model performance, for BERT 3 other hyperparameters are crucial to assess. 

Among these \textit{learning rate}, which sets the speed of updating the weights of the model. Considered values are 1e-3, 1e-4 and 1e-5. As discussed in Chapter~\ref{chap:algorithms}, \textit{number of hidden layers} and \textit{number of attention heads} also play the crucial part in tuning the BERT model, hence this research aims at trying different combinations between these parameters as well. 

After applying these steps and gathering the combinations of window/step sizes, LSTM and BERT models can be directly trained on the resulting data. 

Unlike LSTM and BERT, TFT handles data differently under-the-hood, making it unimportant to enhance train data set. Instead, it is possible to feed only the data from \textit{chartevents} table and specify a \textit{max\_encoder\_length} parameter, which denotes the number of previous records that are taken into account, making it \textit{sliding window}-like approach. Since TFT can handle static data along with time series, admissions table will also be included to enhance the data from \textit{chartevents}.

\chapter{Results}
\section{Classic ML Models on Static Data}
Study showed, that following hyperparameters were optimal for the models:

\begin{itemize}
    \item \textbf{KNN}: 
        \begin{enumerate}
            \item \textbf{Number of neighbors}: 35.
            \item \textbf{Weight}: \textit{distance}.
            \item \textbf{Metric}: \textit{Eucledian}.
        \end{enumerate}
    \item \textbf{SVM}: 
        \begin{enumerate}
            \item \textbf{C}: 0.104.
            \item \textbf{Kernel}: \textit{linear}.
            \item \textbf{Gamma}: \textit{scale}.
            \item \textbf{Degree}: 3.
        \end{enumerate}
    \item \textbf{Random Forest}: 
        \begin{enumerate}
            \item \textbf{Number of estimators}: 943.
            \item \textbf{Maximum depth of trees}: 26.
            \item \textbf{Minimum samples to split}: 5.
            \item \textbf{Minimum samples in a leaf}: 1.
            \item \textbf{Maximum features to consider}: \textit{sqrt}.
        \end{enumerate}
    \item \textbf{XGBoost}: 
        \begin{enumerate}
            \item \textbf{Number of estimators}: 450.
            \item \textbf{Maximum depth of trees}: 13.
            \item \textbf{Learning rate}: 0.0116.
            \item \textbf{Subsample ratio of training instances}: 0.8961.
            \item \textbf{Column subsample by tree}: 0.5711.
            \item \textbf{Minimum loss reduction}: 0.0033.
            \item \textbf{L1 regularization}: 1.5189.
            \item \textbf{L2 regularization}: 1.4634.
            \item \textbf{Minimum child weight}: 4
        \end{enumerate}
        \item \textbf{CatBoost}: 
        \begin{enumerate}
            \item \textbf{Number of boosting iterations}: 1415.
            \item \textbf{Maximum depth of trees}: 6.
            \item \textbf{Learning rate}: 0.1342.
            \item \textbf{L2 regularization}: 1.8871.
            \item \textbf{Number of splits for numerical features}: 145.
            \item \textbf{Subsample temperature}: 0.4268.
            \item \textbf{Random noise applied to features}: 1.7541.
        \end{enumerate}
\end{itemize}

After training models with the hyperparameters (provided above) and specifically selected features during the \textit{Feature Selection} step applied to each model separately, they produced following results:

\begin{center}
    \begin{tabular}{|c|c|c|c|c|}
        \hline
        \textbf{Model} & \textbf{Accuracy} & \textbf{Precision} & \textbf{Recall} & \textbf{F1 Score} \\
        \hline
        \textbf{SVM} & 0.551538 & 0.645051 &  0.482515 & 0.476625 \\
        \textbf{KNN} & 0.651916 & 0.652437 & 0.651916 & 0.640715 \\
        \textbf{Random Forest} & 0.677280 & 0.679044 & 0.677280 & 0.669680 \\
        \textbf{XGBoost} & 0.667026 & 0.667456 & 0.659282
 & 0.661837 \\
        \textbf{CatBoost} & 0.662709 & 0.664003 & 0.662709 & 0.656987 \\         
        \hline
    \end{tabular}
\end{center}
\captionof{table}{Results of Classic ML models}

\vspace{15pt}

According to the results, \textbf{Random Forest} slightly outperformed other models at every metric (\textit{Accuracy}, \textit{Precision}, \textit{Recall} and \textit{F1 Score}), making it the most applicable model in the context of predicting ICU LOS for patients with neurological disorders. Although it is worth noting that the competition from \textbf{CatBoost} and \textbf{XGBoost} was quite close. At the same time there is a distinguishable difference in the performance of \textbf{SVM} and \textbf{KNN}, proving that these models are less applicable. 

\textbf{Random Forest} model highlighted as the most significant features test results of \textit{pH}, \textit{sodium} and \textit{white blood cell}. This correlates with the studies on influence of higher pH value on neurological outcome \cite{oussama2020ph}. At the same time low levels of \textit{sodium} (named \textit{hyponatremia}) are associated with more possiblew in-hospital mortality rate \cite{ha2021hyponatremia}. It is also proved that higher white blood cells count is associated with increased clinical stroke severity \cite{giese2018wbc}.

Interestingly, the most important feature for the final \textbf{XGBoost} model was a \textit{first careunit}, which indicates a type of ICU to which patient was firstly admitted, however out of 6311 \textit{icustays} 2522 of them labeled as \textit{Neuro Intermediate} and 1065 labeled as Neuro SICU. 

Apart from that, model heavily relied on the following test results: \textit{free calcium}, and reassured importance of \textit{sodium} and \textit{pH}. These results prove the significance of their neurological impact. 

It is also proved, that the hypocalcemia leads to deterioration of motor control (\textit{extrapyramidal disorders}) and leads to \textit{basal ganglia calcification}, which also results in movement issues \cite{lee2010fluid}.

\textbf{CatBoost} model, which performed quite closely on the MIMIC-IV data, showed similar features as the most important: \textit{first careunit} and \textit{white blood cells} level are the most essential for the final model. However, the next feature by its importance is \textit{anchor year}. The most plausible reason for that might lie in the fact that the medical technologies and quality of care improved over the years, which is quite likely due to the fact that MIMIC-IV dataset includes data over 14 years, as stated in ~\ref{sec:mimic}. Another possibility is that due to the smaller size of the minorities in age groups \textit{SMOTE} method did not adequately distribute and represent them leading to imbalance. Notably, other features which could potentially contribute to models' bias, such as \textit{gender} or \textit{race} were generalized better according to the feature importance as the didn't play the key role in the predictions. This fact remarks capabilities of the \textit{SMOTE} approach usage in the context of current study and notes that is a powerful tool for data enhancement, especially when bias mitigation is needed.

At the same time \textbf{KNN} also showed that the test results of \textit{pH} play significant role in estimating LOS. Unlike other models, it also used \textit{bicarbonate} results with higher weight. According to \citet{ghauri2019bicarbonate}, reduced \textit{bicarbonate} level resulting in severe metabolic acidosis must be monitored and treated in order to improve survivability in ICUs.

It is also worth noting, that, generally, all models (although not significantly) performed worse on the second class (2-7 days) and better on the third class (more than a week). This might be the result of \textbf{lack of distinct patterns}, which could have helped identify this specific class better. Another potential issue is the distribution of LOS in the \textit{admissions} table:

\begin{figure}[H]
    \centering
    \includegraphics[width=0.8\linewidth]{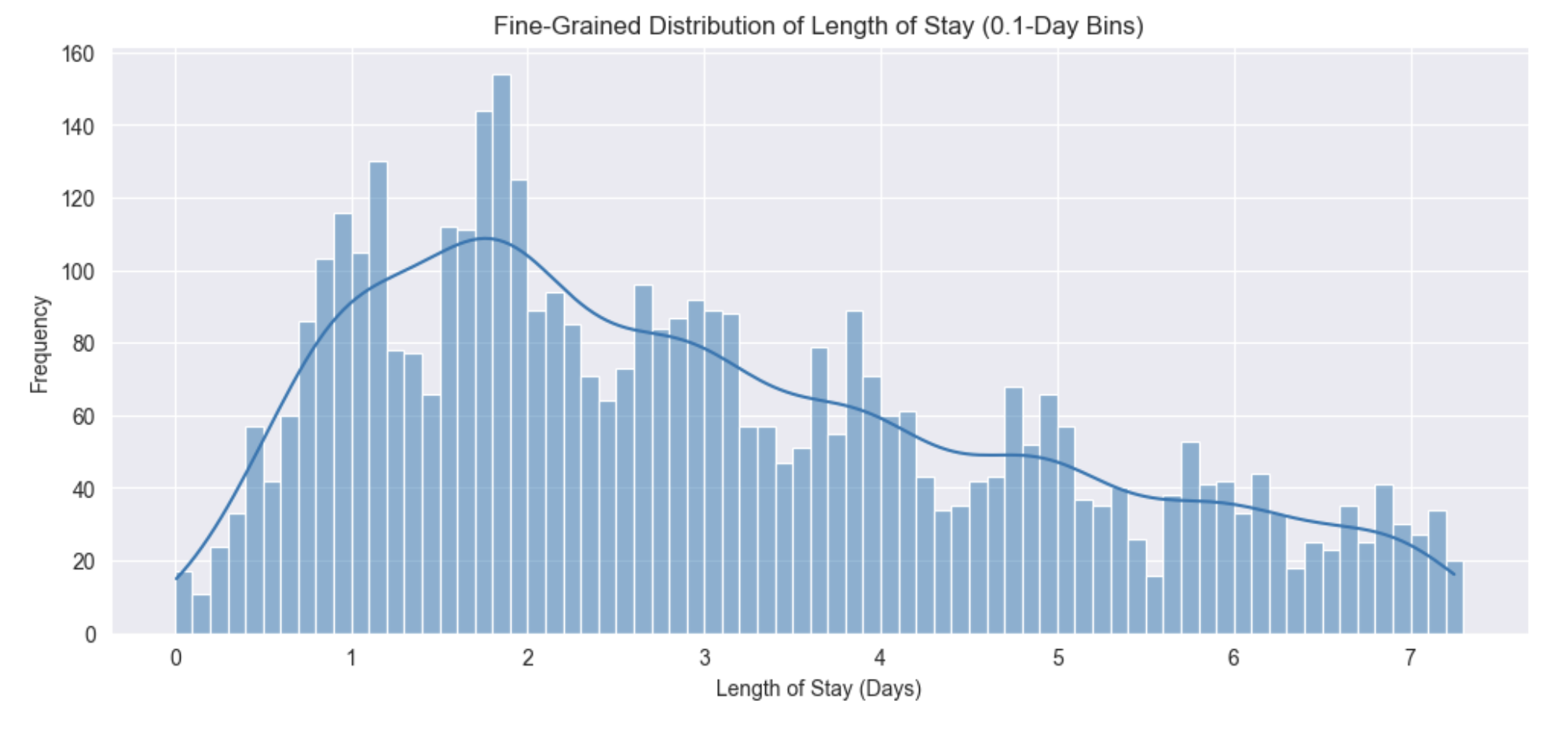}
    \caption{LOS Distribution in Admissions Table}
    \label{fig:distribution_of_los}
\end{figure}

Figure~\ref{fig:distribution_of_los} only displays distribution for those LOS which are less than 7.25 (so that the borders between classes might be analyzed. It also shows that there is a significant amount of patients near the exact border between two classes (2 days).

\section{LSTM and Transformer Models on Time Series Data}

\subsection{LSTM Results}

LSTM model produced following results:

\begin{table}[h]
    \centering
    \begin{tabular}{|c|c|c|c|c|c|}
        \hline
        Window Size & Step Size & Accuracy & Precision & Recall & F1 Score \\
        \hline
        128  & 32  & 0.6021 & 0.5670 & 0.5681 & 0.5613 \\
        128  & 64  & 0.5543 & 0.5094 & 0.5125 & 0.4762 \\
        256  & 32  & 0.6330 & 0.5916 & 0.5933 & 0.5873 \\
        256  & 64  & 0.4890 & 0.4556 & 0.4394 & 0.3829 \\
        256  & 128 & 0.4740 & 0.4188 & 0.4354 & 0.4193 \\
        512  & 32  & 0.6904 & 0.5916 & 0.5933 & 0.5873 \\
        512  & 64  & 0.5826 & 0.5545 & 0.5313 & 0.5151 \\
        512  & 128 & 0.4911 & 0.4422 & 0.4450 & 0.4296 \\
        512  & 256 & 0.4197 & 0.3835 & 0.3838 & 0.3782 \\
        \textbf{1024} & \textbf{32}  & \textbf{0.7204} & \textbf{0.6553} & \textbf{0.6408} & \textbf{0.6338} \\
        1024 & 64  & 0.6124 & 0.5284 & 0.5193 & 0.5174 \\
        1024 & 256 & 0.4878 & 0.4255 & 0.4239 & 0.4224 \\
        \hline
    \end{tabular}
    \caption{Performance Metrics for Different Window and Step Sizes}
    \label{tab:lstm_performance_metrics}
\end{table}

\vspace{15pt}

According to Table \ref{tab:lstm_performance_metrics}, the smaller \textit{step size} (resulting in bigger train (and test) data) improved the performance, regardless of \textit{window size}. Another key takeaway is that bigger \textit{window sizes} accumulate better score. 

\begin{figure}[H]
    \centering
    \includegraphics[width=1\linewidth]{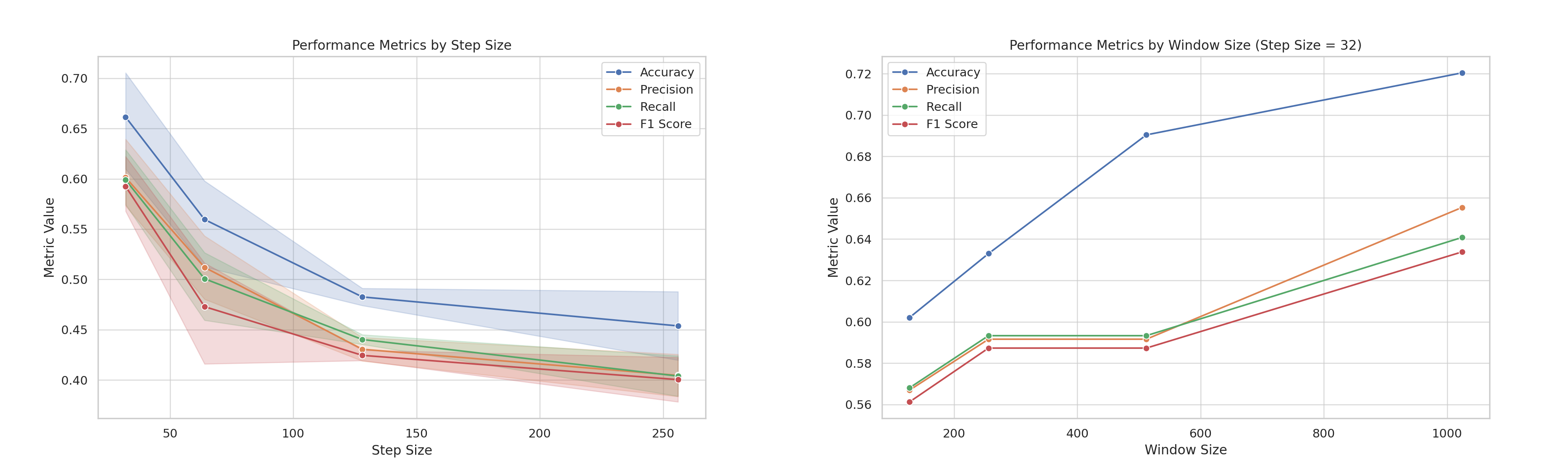}
    \caption{Step Size and Window Size Impact on Metrics}
    \label{fig:performance_metrics}
\end{figure}

Therefore, the best performance of LSTM was on data enhanced by \textit{sliding window} approach with \textit{window size} of \textbf{1024} and \textit{step size} of \textbf{32}. Achieved \textbf{Accuracy} is \textbf{0.7204}, \textbf{Precision} is \textbf{0.6553}, \textbf{Recall} is \textbf{0.6408} and \textbf{F1 Score} is \textbf{0.6338}.

Interestingly, LSTM model also predicted second class worse than others, although the difference was more severe than of classic ML models.

As LSTM does not have feature importance retrival functionality implemented, this study used \textbf{Permutation Importance} approach proposed by \citet{altmann2010pi}. It is used to estimate the contribution of each feature by firstly calculating the model's accuracy (or any other selected metric) on the whole test dataset, and the permute the array of column by iteratively dropping each one of them to assess the difference of the results. According to this approach, an absence of the most significant feature results in higher loss of model's quality. Best LSTM model yielded these tests as the most important during \textit{Permutation Importance} approach:

\begin{table}[h]
    \centering
    \begin{tabular}{|l|c|}
        \hline
        \multicolumn{1}{|c|}{\textbf{Feature}} & \textbf{Importance} \\
        \hline
        Goal Richmond-RAS Scale & 0.023399 \\
        Ventilator Mode (Hamilton) & 0.019704 \\
        Ventilator Mode & 0.018473 \\
        CAM-ICU Inattention & 0.011084 \\
        Daily Wake Up & 0.009852 \\
        Daily Wake Up Deferred & 0.008621 \\
        \hline
    \end{tabular}
    \caption{Feature Importance Scores LSTM}
    \label{tab:feature_importance_lstm}
\end{table}

\textit{Goal Richmond-RAS Scale (RASS} test is used to monitor patients' changes in the sedation levels \cite{woodrow2018rass}. This finding also correlates with the research conducted by \citet{rashidi2020rass}, which showed that the usage of RASS may result in shorter ICU stays. Therefore, this test needs to be closely monitored by the practitioners.

\textit{Ventilator Mode (Hamilton)} and \textit{Ventilation Mode} represent specific settings of the ventilation used in the ICU. The first one specifically displays the mode of Hamilton Ventilation device. In MIMIC-IV dataset, there are 15 different modes of ventilation each signifying the state of the patient. \citet{pelosi2011vent} showed how different conditions require specific ventilation modes. Therefore, knowing in advance the ventilation setup carries important information for predicting LOS in ICU.

\textit{CAM-ICU Inattention} metric is used to track the delirium of patients in ICUs. It is meant to assess whether there is a disturbance of attention in place. It was proved by \citet{dziegielewski2021cam}, that patients with delirium are expecting worse results. Thus, this test emerges as an important one for the current study's main goal. 

\textit{Daily Wake Up} feature displays whether there were interruptions in daily sedation. \textit{Daily Wake Up Deferred} specifies which, if any. \citet{kress2000dwu} showed, that these interruptions result in change of ventilation duration and LOS. Such interruptions happen due to medical considerations or change in patients' condition, which have high impact on LOS.

\subsection{BERT Results}

BERT model, along with the \textit{window size} and \textit{step size} parameters was also tested with changing \textit{hidden layer size} and \textit{attention layer size}. It was estimated in beforehand, that \textbf{learning rate} of \textbf{0.0001} is optimal for the model.

\begin{figure}[H]
    \centering
    \includegraphics[width=1\linewidth]{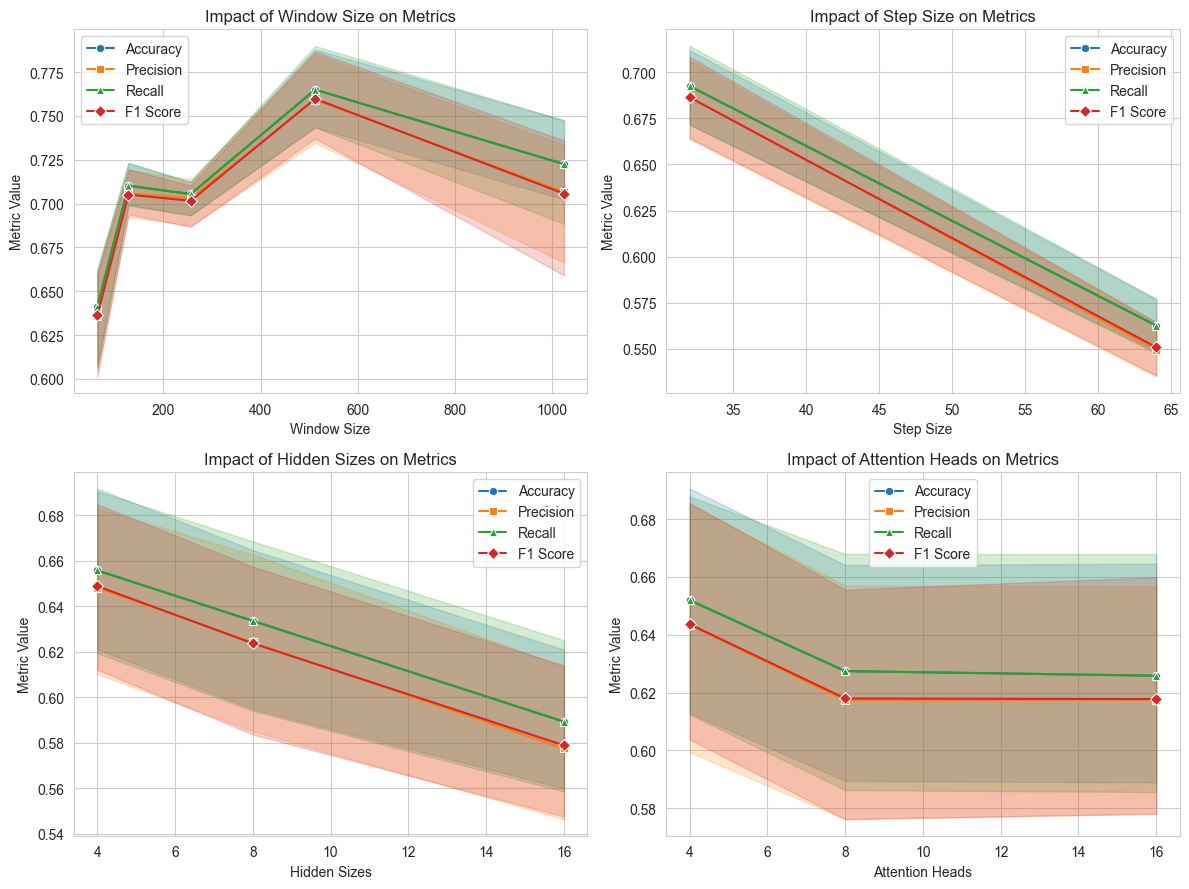}
    \caption{Impact of Parameters on Metrics}
    \label{fig:bert_metrics}
\end{figure}

Figure~\ref{fig:bert_metrics} shows that, unlike LSTM, optimal \textbf{window size} is \textbf{512}. One plausible reason for that might be the \textit{attention mechanism} of the algorithm. More specifically, number of computations increase 4 times while expanding window from 512 previous records to 1024 \cite{li2019mb}. Another explanation might be that with window size of 1024 train dataset consist of lesser amount of data. However, it could also signal that to successfully predict LOS in ICU lesser horizon is needed. Some defining factors might have an impact on limiting time, making older tests obsolete. 

Similarly to LSTM, \textbf{step size} negatively correlates with metrics, meaning that shorter steps are superior. Possible explanation is that it produces more data (roughly 2 times more time series for train data), which results in better quality.

The results also suggest that choosing smaller \textbf{size of hidden layer} also result in better performance. Reason for that might be the computational efficiency (vs. bigger hidden layer) and more sustainable against overfitting. 

Likewise, smaller \textbf{number of attention heads} produced better results. It could be due to the fact that smaller amount of attention heads is slower to overfit and could better focus on global trends rather than on local ones. 

Best results, achieved by BERT model is:

\begin{table}[h]
    \centering
    \begin{tabular}{|c|c|}
        \hline
        \textbf{Metric} & \textbf{Value} \\
        \hline
        Accuracy  & 0.8041 \\
        Precision & 0.8011 \\
        Recall    & 0.8041 \\
        F1-score  & 0.8023 \\
        \hline
    \end{tabular}
    \caption{Best BERT Performance Metrics}
    \label{tab:bert_performance_metrics}
\end{table}

Results in Table~\ref{tab:bert_performance_metrics} were achieved by selecting these hyperparameters: \textbf{window size} of \textbf{512}, \textbf{step size} of \textbf{32}, \textbf{hidden layer size} of \textbf{4}, \textbf{16 attention heads} and \textbf{learning rate} of 0.0001. 

\begin{table}[h]
    \centering
    \begin{tabular}{|l|c|}
        \hline
        \multicolumn{1}{|c|}{\textbf{Feature}} & \textbf{Importance} \\
        \hline
        Ventilator Mode & 0.141203 \\
        Ventilator Mode Hamilton & 0.133630 \\
        Pain Level & 0.097550 \\
        Pain Level Response & 0.082851 \\
        Daily Wake Up Deferred & 0.073051 \\
        \hline
    \end{tabular}
    \caption{Feature Importance Scores BERT}
    \label{tab:bert_feature_importance}
\end{table}

Feature importance calculation for BERT model was done in the same fashion as for the LSTM model, Permutation Importance approach was used. Interestingly, according to Table~\ref{tab:bert_feature_importance}, BERT model identified \textit{Ventilation Modes} and \textit{Daily Ware Up Deferred} as the important ones, similar to the LSTM Model. However, BERT also interpreted \textit{Pain Level} and \textit{Pain Level Response} as the crucial ones as well. While \textit{Pain Level} is patient's self-report test, \textit{Pain Level Response} is used to monitor how well patient respond to the medication. Both of the use Verbal Description Scale (VDS). Because of their importance it is often suggested to minimize sedation so that the patients would be more able to participate in such tests \cite{chanques2022pl}.

Since BERT and LSTM models were employed to solve the same problem in the same context, it is fair to compare them. BERT model outperforms LSTM at each metric by a significant margin, as shown in Table~\ref{tab:comparison_lstm_bert}. 

\begin{table}[h]
    \centering
    \begin{tabular}{|c|c|c|c|c|}
        \hline
        \textbf{Model} & \textbf{Accuracy} & \textbf{Precision} & \textbf{Recall} & \textbf{F1 Score} \\
        \hline
        LSTM & 0.7204 & 0.6553 & 0.6408 & 0.6338 \\
        BERT & 0.8041 & 0.8011 & 0.8041 & 0.8023 \\
        \hline
    \end{tabular}
    \caption{Comparison of Best Results Achieved by LSTM and BERT}
    \label{tab:comparison_lstm_bert}
\end{table}

\subsection{Temporal Fusion Transformer Results}

TFT algorithm was employed as a probable future work for the current study. Its powerful design and adaptability for different domains make it a promising method to approach predicting LOS in ICUs in general. TFT model requires complex setup and rigorous data preparation, while offering significant amount of functionality.

Although the attempt to leverage such algorithm was made during the current research, lack of enough computing resources (only 2 GPUs per job allowed) led to the unsatisfactory results, reaching only the following performance on the selected metrics:

\begin{table}[h]
    \centering
    \begin{tabular}{|c|c|}
        \hline
        \textbf{Metric} & \textbf{Value} \\
        \hline
        Accuracy  & 0.3760 \\
        Precision & 0.3598 \\
        Recall    & 0.3599 \\
        F1-score  & 0.3521 \\
        \hline
    \end{tabular}
    \caption{TFT Performance Metrics}
    \label{tab:tft_performance_metrics}
\end{table}

While the most probable reason might be in irregular time intervals in \textit{chartevents\_original} table or a sparse nature of \textit{chartevents\_by\_minute}, current research was limited in the computational resources and connection to the server time, restraining from train on higher amount of epochs. For comparison, LSTM model used 30 epochs to achieve best results, BERT model used 150 epochs. These models required almost 90 times less time to finish an epoch of training when compared to TFT. This nuance and time constraints significantly affected training TFT. Another issue is higher complexity of computations dut to GRNs (as discussed in \ref{chap:algorithms}) which also  limited both the research and model's capability in that setup.

\chapter{Conclusions}

This study showed that Random Forest, XGBoost and CatBoost algorithms are capable of forecasting LOS for patients with neurology disorders based on the their admission data. This proved that the models based on utilizing Decision Trees are the most suitable for such application.

Another key finding is that predicting only the remaining LOS in ICU based on a window of several events (tests) in the past might result in higher accuracy comparing to the admissions data. Both LSTM- and BERT-based models outperformed Random Forest on the selected metrics. Although, it is worth noting that the context of their usage is different and generally the results should not be compared between distinguishing tasks.

Current research established a strategy to forecast the LOS in ICU for patients with neurological disorders both by data on admission with test results and by chartevents, tracking the condition of patients in ICU. Moreover, with the help of feature importance of the developed models, several important aspects for the practitioners in hospitals are either proved or discovered. Since it is the first approach to specifically target neurology patients in an aim to develop an ML system to predict (remaining) LOS, proposed steps to conduct further research on this topic might be reused and applied in order to build upon current one.

Current study certainly has its limitations. First of all, only one dataset was used, MIMIC-IV, which included data only from one hospital. Next steps might include integration of data from different sources (and different countries as well). This might improve generalization of the developed models.

Secondly, models predicted classes a bit unevenly, despite mitigating that issue with data enhancement technique \textit{SMOTE}. While it did not significantly deteriorate the outcomes, binning still might be improved. In this research borders were defined by (almost) even distribution of classes while maintaining them as integers for better human representation. However, a significant amount of patients in the used dataset had LOS near the border between classes, which resulted in worsened performance. One way to address this issue is to revert problem back from classification to regression.

Lastly, while this research used different approaches and model architectures, there is certainly many more, including previously discussed TFT model. Other examples include hybrid architectures, such as \textit{CNN-LSTM} \cite{lu2020cnnlstm}. Another interesting architecture is diffusion models. There is already one proposed (TimeGrad) by \citet{rasul2021timegrad}, which can handle multivariate probabilistic time forecasting.

\bibliography{references}

\end{document}